\documentclass[10pt,twocolumn,letterpaper]{article}

 \usepackage{cvpr}              
\usepackage{marvosym}
\usepackage{enumitem}
\usepackage{multirow}
\usepackage[accsupp]{accessibility}
\usepackage{amsmath}
\usepackage{pifont}
\usepackage{ulem}
\usepackage{dsfont}
\usepackage{colortbl}
\usepackage[table,xcdraw]{xcolor}
\usepackage{algorithm}
\usepackage{algpseudocode}
\usepackage{multirow}
\definecolor{cvprblue}{rgb}{0.21,0.49,0.74}
\usepackage[pagebackref,breaklinks,colorlinks,allcolors=cvprblue]{hyperref}

\def\paperID{2260} 
\def\confName{CVPR}
\def\confYear{2026}

\title{MSGNav: Unleashing the Power of Multi-modal 3D Scene Graph 

for Zero-Shot Embodied Navigation
}
\author{Xun Huang\textsuperscript{\rm 1,\rm 2,\rm 6} \quad Shijia Zhao\textsuperscript{\rm 1,\rm 2} \quad Yunxiang Wang\textsuperscript{\rm 4} \quad   Xin Lu\textsuperscript{\rm 5,\rm 6} \\ Wanfa Zhang\textsuperscript{\rm 1,\rm 2} \quad Rongsheng Qu\textsuperscript{\rm 3,\rm 6} \quad Weixin Li\textsuperscript{\rm 3,\rm 6}\Letter \quad Yunhong Wang\textsuperscript{\rm 3,\rm 6} \quad Chenglu Wen\textsuperscript{\rm 1,\rm 2}\Letter\\
     \textsuperscript{\rm 1}Fujian Key Laboratory of Urban Intelligent Sensing and Computing, Xiamen University, China \\
    \textsuperscript{\rm 2}Key Laboratory of Multimedia Trusted Perception and Efficient Computing, \\Ministry of Education of China, Xiamen University, China \\ 
    \textsuperscript{\rm 3}Beihang University, China
    \textsuperscript{\rm 4}Nanyang Technological University, Singapore \\
    \textsuperscript{\rm 5}University of Chinese Academy of Sciences, China 
    \textsuperscript{\rm 6}Zhongguancun Academy, China 
\\
\footnotetext{†Corresponding authors.}
}

\begin{document}

\twocolumn[{
\renewcommand\twocolumn[1][]{#1}

\maketitle
\begin{center}
\vspace{-35pt}
\includegraphics[width=1\textwidth]{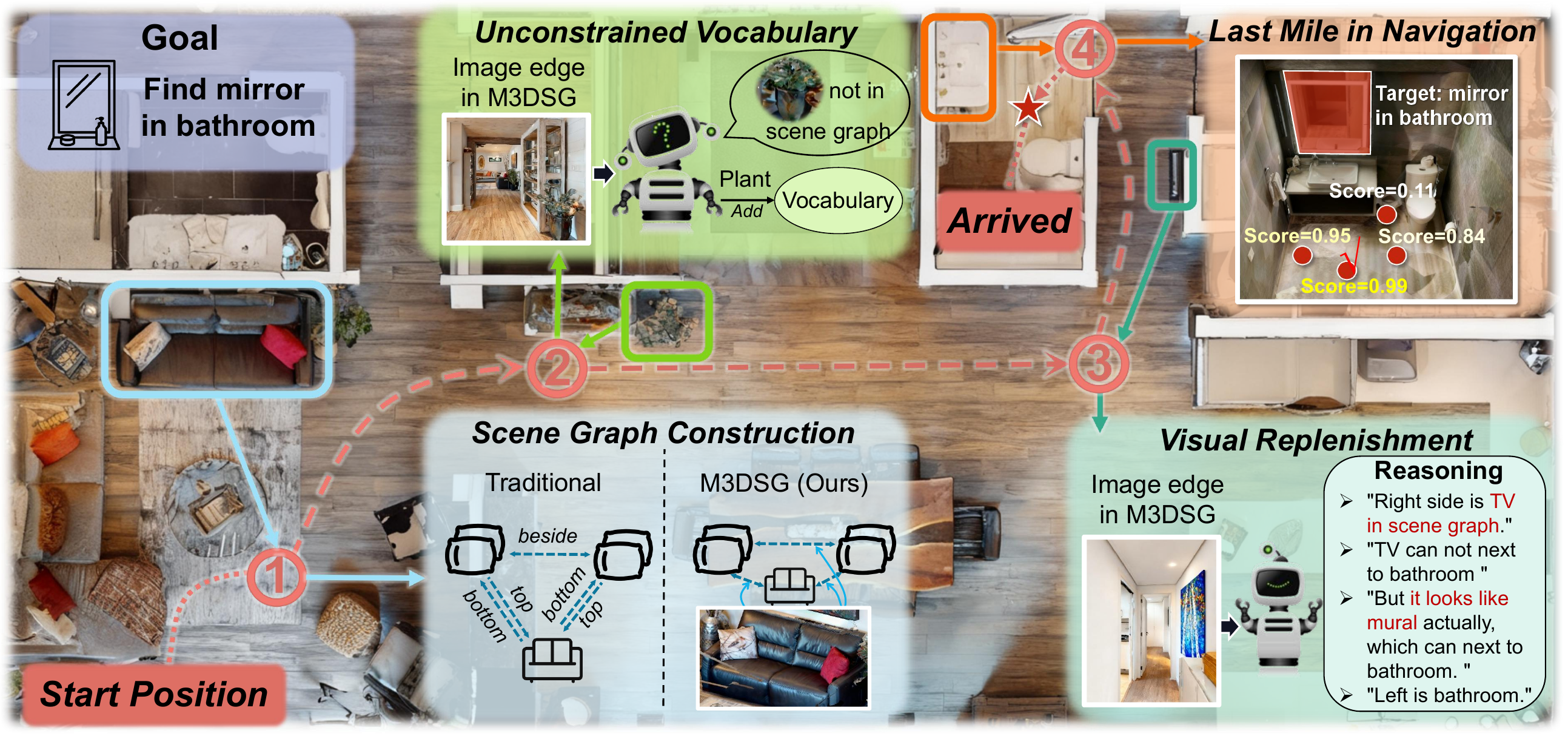}
\centering

\vspace{-5pt}
\captionof{figure}{\textbf{One example illustrating the key insights of our work}. We introduce the Multi-modal 3D Scene Graph (M3DSG) as an alternative to traditional 3D scene graphs, enabling efficient scene graph generation. By incorporating dynamically preserved image-edge information, M3DSG supports unconstrained vocabulary and enhanced visual replenishment for the agent, thereby allowing more comprehensive and context-aware scene understanding in navigation tasks. Furthermore, to address the last-mile problem of selecting the optimal navigation viewpoint given a target location, we propose a visibility-based scoring mechanism for candidate viewpoints.} 
\label{fig:teaser}
\end{center}
}]
\begin{abstract}
Embodied navigation is a fundamental capability for robotic agents operating. Real-world deployment requires open vocabulary generalization and low training overhead, motivating zero-shot methods rather than task-specific RL training. However, existing zero-shot methods that build explicit 3D scene graphs often compress rich visual observations into text-only relations, leading to high construction cost, irreversible loss of visual evidence, and constrained vocabularies.
To address these limitations, we introduce the Multi-modal 3D Scene Graph (M3DSG), which preserves visual cues by replacing textual relational edges with dynamically assigned images. Built on M3DSG, we propose \texttt{MSGNav}, a zero-shot navigation system that includes a Key Subgraph Selection module for efficient reasoning, an Adaptive Vocabulary Update module for open vocabulary support, and a Closed-Loop Reasoning module for accurate exploration reasoning. Additionally, we further identify the “last mile” problem in zero-shot navigation — determining the feasible target location with a suitable final viewpoint, and propose a Visibility-based Viewpoint Decision module to explicitly resolve it. Comprehensive experimental results demonstrate that \texttt{MSGNav} achieves state-of-the-art performance on the challenging GOAT-Bench and HM3D-ObjNav benchmark.  The 
\vspace{-10pt}
code will be publicly available at \href{https://github.com/ylwhxht/MSGNav}{here}.
\end{abstract}    
\section{Introduction}
\label{sec:intro}
\renewcommand{\thefootnote}{} 
\footnotetext{\Letter~Corresponding author.}

Embodied navigation is a fundamental capability of robotic agents, enabling them to accomplish diverse tasks in realistic environments \cite{ovon}. Such tasks are typically goal-oriented, with targets defined by objects, natural language descriptions, or images \cite{unigoal}. Recently, the challenging open-vocabulary embodied navigation task \cite{mtu3d} has gained increasing attention, as real-world agents are required to generalize to a wide and unconstrained range of object categories \cite{ovon}.

Previous RL-based embodied navigation methods suffer from poor generalization and a large sim-to-real gap \cite{sg-nav}. Inspired by rapid advances in pre-trained large model techniques \cite{gpt4o,qwen}, zero-shot embodied navigation approaches have emerged at this historic moment. Zero-shot navigation system does not require any training or fine-tuning before being applied to real-world scenarios. These approaches \cite{3dmem, dynavlm, tango} substantially reduce training time and computational cost while enabling open‑vocabulary navigation, thereby facilitating deployment in real‑world applications.

Notably, a series of methods \cite{3dsgnav1, 3dsgnav2} that build explicit 3D scene graphs from observations and leverage LLMs to drive exploration have shown promising performance on standard goal-oriented navigation benchmarks \cite{hm3d, mp3d}. However, traditional 3D scene graphs overly abstract 3D object relations into simple textual labels (e.g., “top”, “beside”), which seriously impedes progress on zero-shot embodied navigation. As illustrated in Fig. \ref{fig:teaser}, the extreme abstracted 3D object-relation in traditional 3D scene graphs introduces three main limitations:

\begin{itemize}[label=\ding{56}]
    \item \textbf{Expensive Construction}. Traditional 3D scene graphs rely on frequent MLLM queries for relation inference, incurring substantial token and time overhead.
    \item \textbf{Visual Deficiency}. Converting visual observations into text-only graphs discards visual evidence, leading to ambiguity and sensitivity against perception errors.
    \item \textbf{Constrained Vocabulary}. Novel categories beyond a preset vocabulary cannot be represented, limiting generalization in 3D scene graph-based methods.

\end{itemize}

\begin{figure}[!t]
    \includegraphics[width=1.0\columnwidth]{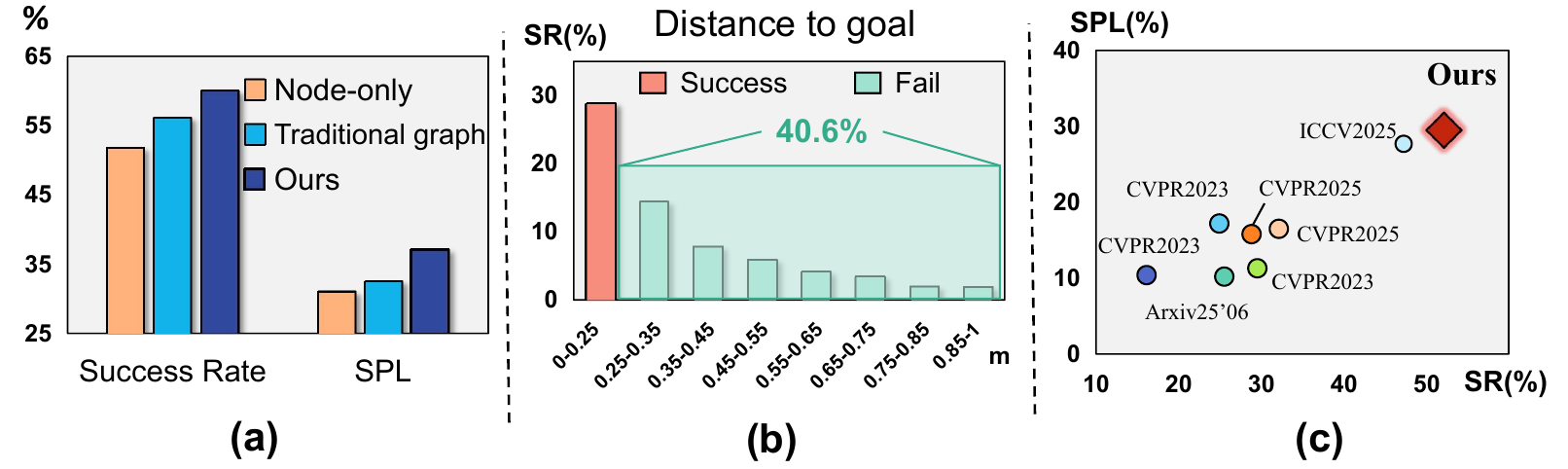}
  \caption{Performance comparisons between our \texttt{MSGNav} and other existing methods for embodied navigation on Goat-Bench \cite{goatbench}: the multi-modal open-vocabulary navigation benchmark. (a) The superiority of our M3DSG over traditional 3D scene graphs. (b) Distance statistics from the goal for the previous method (3D-Mem \cite{3dmem} as an example). (c) Our \texttt{MSGNav} system achieves state-of-the-art performance on the challenging Goat-Bench. }
  \label{intro}
\end{figure}

Based on these observations, we conclude that visual information is indispensable to the real-world navigation process. Therefore, as illustrated in Fig. \ref{fig:teaser}, we introduce the Multi-modal 3D Scene Graph (M3DSG), which replaces the pure-text relational edges with dynamically assigned images to incorporate visual cues, and facilitates efficient and open-vocabulary inference.  
Specifically, our M3DSG effectively addresses the three limitations above:

\begin{itemize}[label=\ding{51}]
    \item \textbf{Efficient Construction}: M3DSG  eliminates costly MLLM queries, greatly reducing construction overhead.
    \item \textbf{Visual Replenishment}: M3DSG provides visual contexts to enhance robustness against perception errors.
    \item \textbf{Unconstrained Vocabulary}: M3DSG utilizes preserved visual contexts to expand vocabulary, enabling open-vocabulary generalization dynamically.
\end{itemize}

As shown in Fig. \ref{intro}(a), our M3DSG markedly outperforms traditional text-only 3D scene graphs in the zero-shot embodied navigation task. Building on M3DSG, we further propose \texttt{MSGNav}, a navigation system that unleashes the power of \textbf{M}ulti-modal 3D \textbf{S}cene \textbf{G}raph for zero-shot embodied \textbf{Nav}igation. As illustrated in Fig. \ref{fig:teaser}, \texttt{MSGNav} first selects target‑relevant subgraphs from the complex multi-modal 3D scene graph by a key‑subgraph selection module. During the exploration process, \texttt{MSGNav} updates the vocabulary using preserved visual information and incorporates a closed-loop reasoning module for accurate exploration reasoning, thereby enabling unconstrained vocabulary and reliable navigation.

\renewcommand{\thefootnote}{1} 

Additionally, we realize that the \textit{last‑mile} problem in navigation tasks — knowing the target’s location is not equivalent to determining a suitable navigation pose\footnote{\textit{Cannot see the wood for the trees}. Being too close or occluded prevents a clear observation of the target, which is not a good viewpoint.}. Existing zero-shot embodied navigation methods usually stop before that final step. As shown in Fig. \ref{intro}(b), the existing 3D-Mem method \cite{3dmem} successfully approaches the target but still fails in many navigation tasks due to unacceptable viewpoints (distances ranging from 0.25 m to 1.0 m). To address this issue, we introduce a visibility‑based viewpoint decision module in our \texttt{MSGNav}.

As shown in Fig. \ref{intro}(c), \texttt{MSGNav} significantly outperforms state-of-the-art RL‑based and zero-shot methods on the challenging embodied navigation GOAT‑Bench \cite{goatbench}, demonstrating consistent and substantial gains in zero‑shot challenging embodied navigation. Our contributions can be summarized as follows:
\begin{itemize}
    \item \textbf{M3DSG}: We propose a multi-modal 3D scene graph that incorporates visual information, overcoming pure-text limitations and enhancing open-vocabulary scene representation for embodied navigation.
    \item \textbf{Last-Mile}: We identify the last-mile problem
    and propose a visibility‑based viewpoint decision module to address it.
    \item \textbf{MSGNav}: We develop a zero-shot embodied navigation system built on M3DSG, enabling active exploration and achieving state-of-the-art performance on the GOAT-Bench and HM3D-ObjNav embodied navigation datasets.

\end{itemize}

\section{Related Work}

\subsection{Graph-based Scene Exploration}
Unlike traditional feature representations \cite{l4dr, v2xr, srkd}, graph representations \cite{Conceptgraphs,related_14} provide explicit spatial–semantic structures for scene reasoning and integrate well with LLMs/VLMs. Examples include SayPlan \cite{related_32} for 3D scene graph planning, OVSG \cite{related_05} for open-vocabulary grounding, imaginative world modeling \cite{hu2025imaginative}, and BEV graph policies \cite{luo2025learning}. Graph reasoning has also been applied in navigation, e.g. SG-Nav \cite{sg-nav} for hierarchical prompting and graph-retained adaptation \cite{hong2025general} for cross-domain VLN. Yet, the overly abstract textual edges of current scene graphs limit the comprehensive scene context understanding, motivating our multi-modal 3D scene graph (M3DSG).

\subsection{Zero-Shot Navigation}
 Zero-shot navigation allows agents to act in unseen environments without task-specific training, unlike supervised methods \cite{related_31,RL} which require large-scale simulation. Existing zero-shot navigation methods span object-goal (ON), image-goal (IIN), and text-goal (TN). ON methods include CoW \cite{related_12} with CLIP \cite{clip} frontier exploration, ESC \cite{esc}, OpenFMNav \cite{Openfmnav}, VLFM \cite{vlfm} using LLM reasoning, UniGoal \cite{unigoal} for multi-modal unification, and RATE-Nav \cite{li2025rate} with region-aware termination. IIN methods range from Mod-IIN \cite{imggoal2} to SIGN \cite{yan2025sign} for safety-aware drone navigation. TN methods such as InstructNav \cite{Instructnav} generalize instruction following across ON, DDN, and VLN \cite{related_09,related_20,related_26,related_47}. However, most zero-shot work ignores the last-mile challenge of choosing an optimal viewpoint after target localization.

\section{Approach}
In this section, we formalize the zero-shot goal-oriented embodied navigation problem and provide an overview of our pipeline (Sec. \ref{sec:3.1}). We then detail the construction of our Multi‑modal 3D Scene Graph (M3DSG) (Sec. \ref{sec:3.2}), and describe \texttt{MSGNav}, the navigation system that performs effective and closed-loop reasoning over this graph (Sec. \ref{sec:3.3}).

\begin{figure*}[!t]
\centering
    \includegraphics[width=0.85\textwidth]{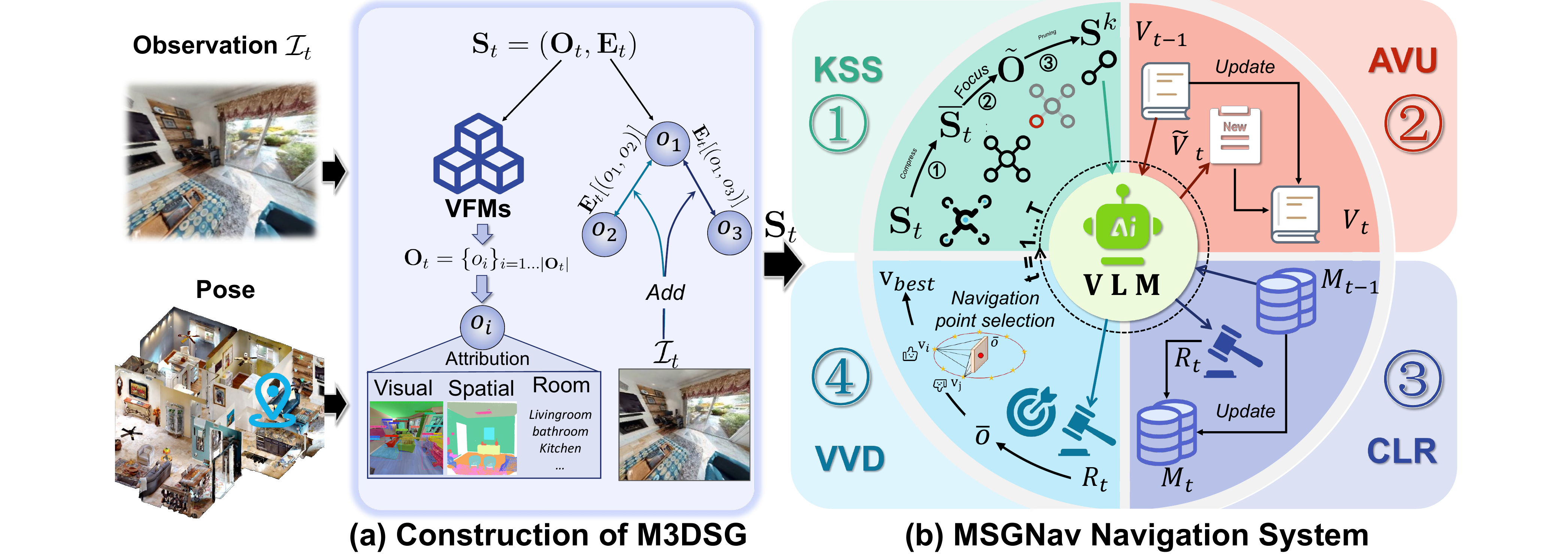}
  \caption{\textbf{The overall framework of our \texttt{MSGNav}}. At time step $t$, the agent incrementally constructs the scene graph $S_t$ based on received observation $\mathcal{I}_t$ and its own pose. $S_t$ includes a set of objects $\textbf{O}_t$ with attributes, namely visual, spatial, and room properties, along with a set of image edges $\textbf{E}_t$ representing object relationships. Subsequently, $S_t$ is processed through KSS, AVU, and CLR modules, before being input to VLM query to obtain the target object $\bar{o}$. Finally, VVD module selects the insightful viewpoint $\textbf{v}_{best}$ as a navigation point.}
  \label{frame}
\end{figure*}

\subsection{Zero‑shot Goal‑Oriented Embodied Navigation} 
\label{sec:3.1}
\subsubsection{Problem definition}
We consider a mobile embodied agent operating in an unknown or partially explored environment (\textit{i.e.} the  lifelong navigation setting). The agent receives target information in the form of an category, natural language description, or reference image. At each time step \( t \), it obtains an RGB-D observation \(\mathcal{I}_t\) and executes an action \(\mathcal{A}_t\) (camera rotation or ego-motion) to actively explore until locating the target. The task is successful if the agent reaches any target viewpoint within \( d \) meters in at most \( T \) steps; otherwise, it fails. In the zero-shot navigation setting, the agent performs navigation without task-specific training or fine-tuning in simulation, thereby reducing training cost and improving generalization to unseen environments and targets.

\subsubsection{Overview}
Zero-shot goal-oriented embodied navigation remains highly challenging for conventional learning-based methods. 
Following current best practices, we utilize off-the-shelf pre-trained models: Vision Foundation Models (VFMs) and Vision Language Models (VLMs) for primitive reasoning. 
While these models excel in visual understanding, they are limited in 3D spatial reasoning. 
To address this gap, we propose the M3DSG, an explicit and multimodal 3D scene graph that is incrementally built to encode scene contexts while preserving the visual evidence. 
Based on M3DSG, the navigation system \texttt{MSGNav} completes goal-oriented navigation with a Visibility-based Viewpoint Decision (VVD) module. 

Formally, at each time step $t$, M3DSG incrementally integrates the RGB-D observation $\mathcal{I}_t$ into an evolving scene graph $\mathbf{S}$. 
\texttt{MSGNav} then uses $\mathbf{S}$ as a prompt to guide VLMs in locating candidate target $\bar{o}$ positions or unexplored frontier for exploration. 
Finally, the VVD module selects an optimal navigation viewpoint $\textbf{v}_{best}$ with the best line of sight to $\bar{o}$, enabling a successful final approach to the target.

\subsection{M3DSG: Multi‑modal 3D Scene Graph} 
\label{sec:3.2}
Traditional 3D scene graphs represent object relations with text-only edges, which are overly abstract and hinder VLMs from accurately understanding scene contexts. 
Inspired by 3D-Mem~\cite{3dmem}, which emphasizes the value of raw images for navigation, we retain visual information when constructing an object-centric 3D scene graph. 
Unlike traditional 3D scene graph~\cite{Conceptgraphs} which uses textual relation edges, our method stores images to describe detailed object relations directly. This image edge preserves the benefits of 3D scene graphs while avoiding repeated, costly model queries and providing a more holistic scene representation. 

To mitigate the inference cost of numerous images during exploration, we design a dynamic allocation algorithm that efficiently converts the multimodal 3D scene graph into prompts, thereby requiring on average only \textbf{4 images} per query to represent the target-related scene context.

\subsubsection{M3DSG Structure Definition}
During the environment exploring process, M3DSG constructs a scene graph $\mathbf{S} = (\mathbf{O}, \mathbf{E})$, 
where $\mathbf{O} = \{ o_i \}_{i=1}^{N_o}$ and $\mathbf{E} = \{ \mathbf{e}_j \}_{j=1}^{N_e}$ denote the sets of objects and edges, respectively. 
Here, $N_o$ is the number of detected objects, and 
$
N_e = \sum_{1 \le x < y \le N_o} \mathds{1}_{\|Pos_{o_x} - Pos_{o_y}\| < \theta}
$
is the number of object pairs $(o_x, o_y)$ whose spatial distance is below $\theta$. 
Each object $o_i$ is described by attributes including corresponding category, 3D coordinates, and room locations. 
Each edge $\mathbf{e}_j$ stores a set of RGB-D images $\mathbf{I}_j = \{\mathcal{I}_k\}_{k=1}^{|\mathbf{I}_j|}$ that represent the pair of corresponding objects to record the visual cues.

\subsubsection{Incremental Construction of M3DSG}
Following prior work~\cite{sg-nav, unigoal}, the scene graph $\mathbf{S}$ is incrementally updated during exploration. 
At step $t$, $\mathbf{S}_{t-1}$ is updated to $\mathbf{S}_t$ with the current RGB-D observation $\mathcal{I}_t$ as:
\begin{equation}
    \mathbf{S}_{t} = \mathcal{M}(\mathbf{S}_{t-1}, \mathcal{I}_t), \quad t \in [1, T], \quad \mathbf{S}_0 = (\varnothing, \varnothing),
\end{equation}
where $\mathcal{M}$ is the graph update function consisting of Object $\mathbf{O}_t$ Update and Edge $\mathbf{E}_t$ Update.

\noindent \textbf{Object Update}. For the RGB-D observation $\mathcal{I}_t$ at time $t$, we extract the frame-level object set 
$
\mathbf{O}^{frame}_{t} = \{o_i\}_{i=1}^{|\mathbf{O}^{frame}_{t}|}
$
by VFMs: YOLO-W~\cite{yolo} for open-vocabulary detection, SAM~\cite{sam} for instance masks, and CLIP~\cite{clip} for visual embeddings. 
Each object $o_i$ is represented as 
$
o_i = \{\mathcal{ID}, \mathcal{C}, \mathcal{P}, \mathcal{B}, \mathcal{M}, \mathcal{PC}, \mathcal{V}, \mathcal{R}\}_i,
$
denoting unique ID, category, 3D position, bounding box, semantic mask, point cloud, visual feature, and room location, respectively. The global object set $\mathbf{O}_t$ is incrementally updated from the previous set $\mathbf{O}_{t-1}$ and the frame-level object set $\mathbf{O}^{frame}_{t}$ as:
\begin{equation}
\mathbf{O}_t = \Phi_{\mathrm{merge}}\big( \Phi_{\mathrm{match}}(\mathbf{O}^{frame}_{t}, \mathbf{O}_{t-1}) \big) \cup \mathbf{O}^{frame}_{t},
\end{equation}
where $\Phi_{\mathrm{match}}$ aligns objects in $\mathbf{O}^{frame}_{t}$ and $\mathbf{O}_{t-1}$ via spatial similarity $(\mathcal{B}, \mathcal{PC})$ and visual similarity $(\mathcal{C}, \mathcal{V})$, 
and $\Phi_{\mathrm{merge}}$ merges features of matched objects~\cite{Conceptgraphs}. 
This process maintains a global object set $\mathbf{O}_t$ within $\mathbf{S}_t$ step by step.

\noindent \textbf{Edge Update}. The core of an edge is a set of images $\mathbf{I}$ in which an object pair co-occurs, indexed by the unique IDs of the object pair, \textit{i.e.}
$
\mathbf{E} : (\mathcal{ID}, \mathcal{ID}) \rightarrow \mathbf{I}.
$
Meanwhile, we maintain a mapping from each image to its corresponding object pair, \textit{i.e.}
$
\mathbf{H} : \mathcal{I} \rightarrow (\mathcal{ID}, \mathcal{ID}).
$ For the current RGB-D frame $\mathcal{I}_t$ at step $t$ with object set $\mathbf{O}^{frame}_{t}$ (from Object Update), 
we generate co-occurring object pairs $(o_x, o_y)$ that appear together in $\mathcal{I}_t$ as:
\begin{footnotesize}
\begin{equation}
\{ (o_x, o_y) \in \mathbf{O}^{frame}_{t} \times \mathbf{O}^{frame}_{t} \mid o_x \neq o_y \ \land \ \|\mathcal{P}_{o_x} - \mathcal{P}_{o_y}\| \leq \theta \},
\end{equation}
\end{footnotesize}
where $\theta$ is the adjacency distance threshold. 
The edge set $\mathbf{E}_t$ is then updated by retrieving the unique IDs $(\mathcal{ID}_{o_x}, \mathcal{ID}_{o_y})$ of each detected pair and appending $\mathcal{I}_t$ to the corresponding image set as:
\vspace{-1mm}
\begin{footnotesize}
\begin{equation}
\begin{aligned}
        \mathbf{E}_t[(\mathcal{ID}_{o_x},\mathcal{ID}_{o_y})]\leftarrow &
\begin{cases}
\{\mathcal{I}_t\} \hspace{2em}  \text{if } (\mathcal{ID}_{o_x},\mathcal{ID}_{o_y}) \notin \mathrm{dom}(\mathbf{E}_{t-1}) \\
\mathbf{E}_{t-1}[(\mathcal{ID}_{o_x},\mathcal{ID}_{o_y})] \cup \{\mathcal{I}_t\} \hspace{1em} \text{otherwise}
\end{cases}, 
\\
    \mathbf{H}[\mathcal{I}_t] \leftarrow &
    \begin{cases}
    \{(\mathcal{ID}_{o_x},\mathcal{ID}_{o_y})\} \hspace{3em}  \text{if } \mathcal{I}_t \notin \mathrm{dom}( \mathbf{H}) \\
\mathbf{H}[\mathcal{I}_t]  \cup (\mathcal{ID}_{o_x},\mathcal{ID}_{o_y}). \hspace{1em}\text{otherwise}
    \end{cases},\\
  &  \mathrm{dom}(X) = \{k \mid \exists v : (k, v) \in X\}.
\end{aligned}
\end{equation}
\end{footnotesize}

This edge update process is efficient, eliminating the need for costly VLM queries. 
Moreover, the stored image recorded in $\mathbf{E}_t$ provides compact and visually grounded representations of object relationships, 
which can be directly utilized for downstream exploration reasoning.

\subsection{\texttt{MSGNav} Embodied Navigation System} 
\label{sec:3.3}
M3DSG provides comprehensive scene representations that preserve rich visual information.
To fully exploit this, we propose the navigation system \texttt{MSGNav}.
As shown in Fig. ~\ref{frame}(b), our \texttt{MSGNav} first significantly reduces the tokens and time cost required for inference by selecting key subgraphs (Sec.~\ref{sec:3.3.1}). 
\texttt{MSGNav} then addresses the out-of-vocabulary issue inherent to pre-set vocabulary by adaptively updating vocabulary with visual evidence from M3DSG (Sec.~\ref{sec:3.3.2}).
Meanwhile, \texttt{MSGNav} performs closed-loop reasoning via decision memory and feedback (Sec.~\ref{sec:3.3.3}). 
Finally, \texttt{MSGNav} tackles the ``\textit{last-mile}'' problem by optimizing navigation points through visibility-based viewpoint decisions (Sec.~\ref{sec:3.3.4}).

\subsubsection{Key Subgraph Selection (KSS)} 
\label{sec:3.3.1}
During exploration, the 3D scene graph $\mathbf{S}$ grows progressively and can become voluminous, thereby impairing the decision-making efficiency \cite{3dmem}. In embodied navigation, however, typically only a small related subgraph to the required target is meaningful. Therefore, processing the full scene graph consumes substantial computational resources without commensurate benefits. 
The key to addressing this issue is to preserve maximal relevant information with the minimal scene graph.  \texttt{MGSNav} extracts the target‑relevant subgraphs via a \textbf{Compress–Focus–Prune} procedure:

\begin{algorithm}[!t]
\caption{Greedy Dynamic Allocation for Pruning}
\label{alg:edge_select}
\begin{algorithmic}[1]
\Require 
    $\text{Image–Object Association Map: } \mathbf{H}$
  $\text{Scene Graph:   } \mathbf{S} = (\mathbf{O},\mathbf{E}), \text{Related Objects: }  \textbf{O}^{rel}$

\State $\textbf{E}^{k} \gets \varnothing$ \Comment{Key edges}
\State $\textbf{O}^{k} \gets \textbf{O}^{rel}$ \Comment{Key objects}
\State $\textbf{U} \gets \varnothing$ \Comment{Uncovered relationship}
\State $\textbf{I}^{total} \gets \varnothing$ \Comment{Total images}
\For{$ o\in \mathbf{O}$} \Comment{Filter relevant edges}
    \If{$\exists~{o}^r \in \textbf{O}^{rel}$: $(\mathcal{ID}_o,\, \mathcal{ID}_{{o}^r}) \in \mathrm{dom}(\mathbf{E})$}
      \State $\textbf{O}^{k} \gets \textbf{O}^{k} \cup o$
      \State $\textbf{U} \gets \textbf{U} \cup (\mathcal{ID}_{o},\mathcal{ID}_{{o}^r})$
      \State $\textbf{I}^{total}  \gets \textbf{I}^{total} \cup \mathbf{E}[(\mathcal{ID}_o,\, \mathcal{ID}_{{o}^r})]$ 
    \EndIf
\EndFor

\While{$\textbf{U} \neq \varnothing$} \Comment{Greedy selection}
  \State $\mathcal{I}^\star \gets \arg\max_{\mathcal{I} \in \textbf{I}^{total}} \ |\mathbf{H}[\mathcal{I}] \cap \mathbf{U}|$
\State $\mathbf{E}^k[\mathbf{H}[\mathcal{I}^\star] \cap \mathbf{U}] \gets \mathcal{I}^\star$
  \Comment{$\mathcal{I}^\star$ cover the most edges.}
  \State $\textbf{U} \gets \textbf{U} \setminus \mathbf{H}[\mathcal{I}^\star]$ \Comment{Remove covered edges}
\EndWhile
\State \Return $\textbf{S}^{k}=(\textbf{O}^{k},\textbf{E}^{k})$   \Comment{Key scene graph}
\end{algorithmic}
\end{algorithm}

\begin{itemize}
    \item \textbf{Compress}. We first simplify the rich but vast  scene graph $\textbf{S}$ into the compact adjacency list representation $\hat{\textbf{S}}=(\hat{\textbf{O}},\hat{\textbf{E}})$. For each object $\hat{o}_i\in\hat{\textbf{O}}$, we retain only essential attributes (ID and category): $\hat{o}_i=\{\mathcal{ID},\mathcal{C}\}_{i}$. The edge set is represented as undirected adjacency list: $\hat{\textbf{E}}_{\hat{o}_i} = \{ID_{\hat{o}_j} | \hat{o}_j \in \hat{\textbf{O}}, (ID_{\hat{o}_i}, ID_{\hat{o}_j}) \in dom(E)\}$.
    \item  \textbf{Focus}. The compressed scene graph $\hat{\textbf{S}}$ significantly reduces the overhead. Therefore, we directly fed it to the VLM to select top‑k related objects $\textbf{O}^{rel} \subset \textbf{O}$ which are the most relevant to the current navigation target.
    \item \textbf{Pruning}. Finally, we construct the key subgraph $\textbf{S}^{k}=(\textbf{O}^{k},\textbf{E}^{k})$ by progressively selecting the image that covers the most edges through a dynamic greedy allocation algorithm (in Algorithm ~\ref{alg:edge_select}).

\end{itemize}

The selected key subgraph $\textbf{S}^{k}$ is used to query the VLM for exploratory reasoning with low cost (on average $\sim$4 images per query):
\begin{equation}
\label{eq_vlm}
    \mathcal{R}_t = \mathrm{VLM}(\textbf{S}^{k}, \mathbf{F}, g, t),
\end{equation}
where $\mathbf{F}$ denotes frontier images for exploration as in ~\cite{3dmem}, 
$g$ is the navigation target, 
and $\mathcal{R}_t$ denotes the selected target or frontier for task response. 
The specific prompt design is provided in the supplementary material.

This technique reduces the token cost by over 95\% while feeding the information to VLM for reasoning, retaining most of the important objects relevant to the task. Consequently, the agent can efficiently focus on task‑relevant entities, improving both navigation speed and performance.

\subsubsection{Adaptive Vocabulary Update (AVU)} 
\label{sec:3.3.2}
Existing methods achieve open-vocabulary navigation simply by  using detectors, e.g., YOLO-W~\cite{yolo} and GroundingDINO~\cite{groundingdino}, 
but remain constrained by preset vocabulary (e.g., ScanNet-200~\cite{scannet}, HM3D~\cite{hm3d}). 
Such constraints hinder VLMs from handling diverse out-of-vocabulary scene representations in the real world. 
Additionally, preset vocabulary is difficult to accurately describe the image target.

To address this issue, \texttt{MSGNav} integrates adaptive vocabulary updates with preserved visual evidence during the exploratory reasoning process.
We first initialize the vocabulary $V_{t=0}$ with the ScanNet-200.
During exploration at time $t$, the VLM inspects edge images $\mathbf{E}_t$ and compares objects in $\mathbf{O}_t$ to propose required novel vocabulary $\hat{V}_t$ (based on the extension of Eq. \ref{eq_vlm}):
\begin{equation}
\label{eq_vlm2}
    \mathcal{R}_t, \hat{V}_t = \mathrm{VLM}(\mathbf{S}^{k}, \mathbf{F}, g, t).
\end{equation}
Then the novel vocabulary $\hat{V}$ will be incorporated into $V_t$:
$V_t = V_{t-1} \cup \hat{V}_t$.
By continuously updating vocabulary $V_t$ in this manner, the scene graph progressively captures richer and more accurate object descriptions, 
supporting navigation beyond preset vocabulary limitations.

\subsubsection{Closed-Loop Reasoning (CLR)} 

\label{sec:3.3.3}
Memory is critical for navigation agents. 
While prior work focuses mainly on scene memory, other forms of memory can also influence decision-making. 
In addition to modeling the scene as perception memory, we introduce the \emph{decision memory} $\mathbf{M}$ for closed-loop reasoning. 
At time step $t-1$, the exploration response $\mathcal{R}_{t-1}$ is stored in a historical action repository $\mathbf{M}_t$ as $
\mathbf{M}_t = \mathbf{M}_{t-1} \cup \mathcal{R}_{t}, \mathbf{M}_0 = \varnothing
$
. Then $\mathbf{M}_t$ is utilized at time step $t$ to assist subsequent decisions (based on the extension of Eq. \ref{eq_vlm2}) as:
\begin{equation}
    \mathcal{R}_t, \hat{V}_t = \mathrm{VLM}(\mathbf{S}^{k}, \mathbf{M}_t, \mathbf{F}, g, t).
\end{equation}

This closed-loop reasoning enhances the accuracy of VLM strategies by memorizing past decision feedback to inform current decisions.

\subsubsection{Visibility-based Viewpoint Decision (VVD)} 
\label{sec:3.3.4}
\begin{figure}[!t]
\centering
    \includegraphics[width=0.87\columnwidth]{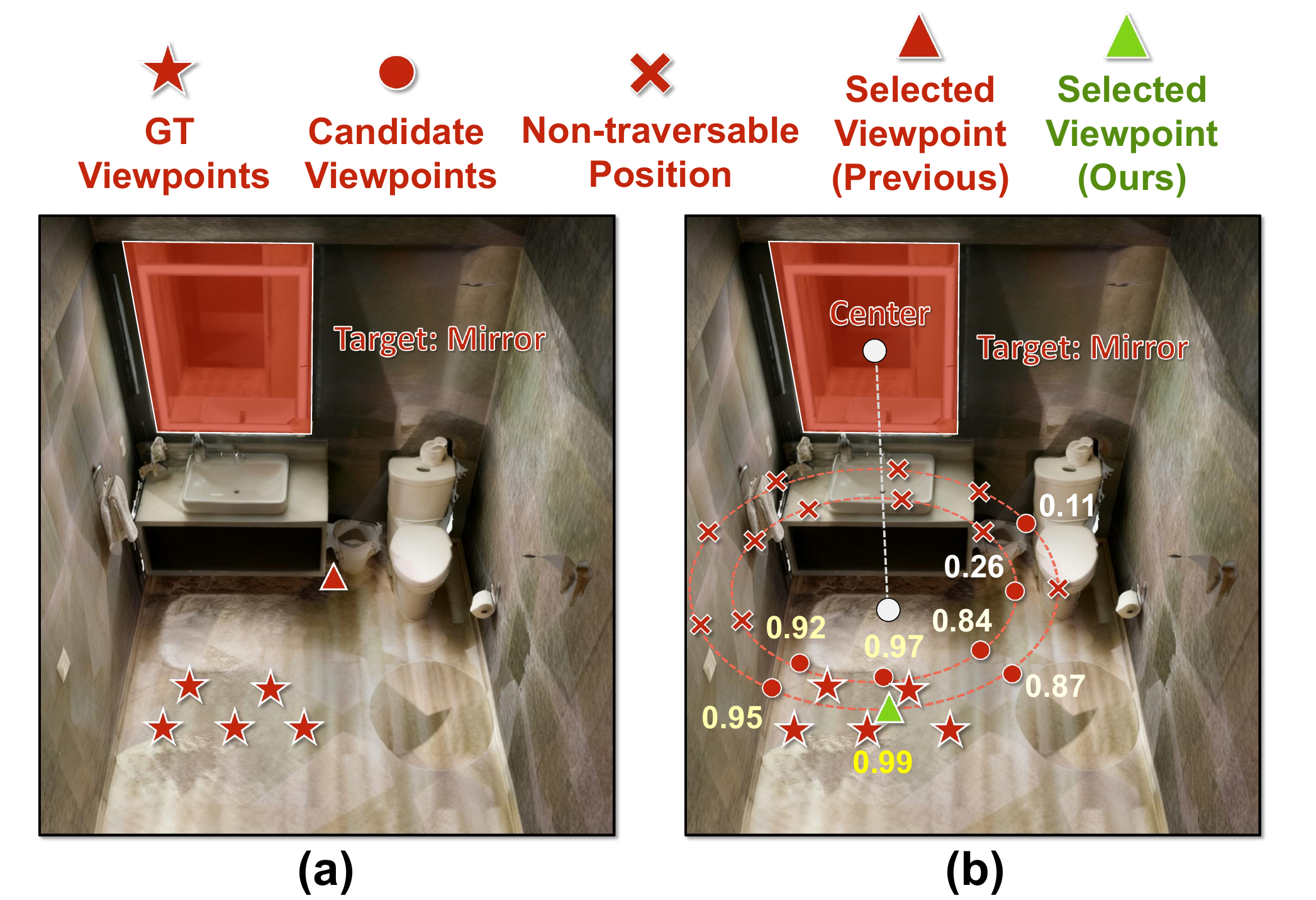}
  \caption{Demonstration of the ``last-mile'' problem. 
(a) Previous methods select the nearest traversable position after target localization, and often fail due to poor viewpoints. (b) Our VVD samples candidate viewpoints and computes visibility, which can select a suitable viewpoint close to GT for successful navigation.}
  \label{lastmile}
\end{figure}
For a navigation agent, the high-level decision process involves the location of the target, while the low-level planner converts it into executable actions. 
Therefore, an optimal final decision should choose a viewpoint with clear visibility to facilitate subsequent operations \cite{wu2025momanipvla, wu2025moto}. 
Previous zero-shot methods \cite{vlmnav, dynavlm, 3dmem, sg-nav} often assume that knowing the target location is enough, and select the nearest traversable position as the navigation viewpoint. 
However, as shown in Fig.~\ref{lastmile}(a), this may yield viewpoints with poor visibility, causing failure even when the target $\bar{o}$ is correctly localized.

We term this the ``\textit{last-mile}'' problem of navigation—identifying a suitable viewpoint despite correctly locating the target $\bar{o}$. 
To solve this problem, we realize that a good viewpoint should have clear visibility of the target. 
To achieve this goal, we propose a \emph{Visibility-based Viewpoint Decision} (VVD) module (in Algorithm~\ref{pseudo:VVD}). As shown in Fig.~\ref{lastmile}(b), VVD uniformly samples candidate viewpoints around target $\bar{o}$ within radii $\mathbf{R} = \{r_j\}_{j=1}^{N_R}$.
For each candidate $\mathbf{v}_i$, VVD computes a visibility score by evaluating occlusion between $\mathbf{v}_i$ and target point cloud $\mathcal{PC}_{\bar{o}}$. 
The viewpoint with the highest visibility score $\textbf{v}_{best}$ is selected as the navigation coordinate. 
Compared with traditional choices, VVD yields viewpoints with superior target visibility.

\begin{algorithm}[!h]
\caption{Visibility-based Viewpoint Decision}\label{pseudo:VVD}
\begin{algorithmic}[1]
    \small
    \Require Target object $\bar{o}$, Scene point cloud $\mathcal{PC}$, Radii set $\mathbf{R}$, Samples number $K$, Camera height $h$, Obstruction distance $\tau$
    \State $\mathbf{c}_{\bar{o}} \gets \frac{1}{|\mathcal{PC}_{\bar{o}}|}\sum_{\mathbf{p}\in\mathcal{PC}_{\bar{o}}}\mathbf{p}$ \Comment{Target center}
    \State $\Theta \gets \left\{ \frac{2\pi k}{K} \,\middle|\, k = 0, \ldots, K{-}1 \right\}$ \Comment{Sampling angles}
    \State $\mathbf{V}_c \gets \bigcup_{r_i \in \mathbf{R}} \left\{ \mathbf{c}_{\bar{o}} + [r_i\cos\theta, h - \mathbf{c}_{\bar{o}}[1], r_i\sin\theta] \mid \theta \in \Theta \right\}$ 

    \State $\mathbf{V}_c \gets \{\mathbf{v} \in \mathbf{V}_c \mid \mathbf{v} \text{ is traversable}\}$ \Comment{Candidate viewpoints}
    \State $S_{best} \gets 0$, $\mathbf{v}_{best} \gets \mathbf{0}$ \Comment{Initialization}
    \For{each $\mathbf{v}_i \in \mathbf{V}_c$}
        \State $\mathcal{Q}({\mathbf{v}_i,\mathbf{p}}) = \left\{ \mathbf{v}_i + t \cdot \frac{\mathbf{p} - \mathbf{v}_i}{\|\mathbf{p} - \mathbf{v}_i\|} \,\middle|\, t \in [\tau, \|\mathbf{p} - \mathbf{v}_i\|-\tau] \right\}$
        \State $\mathcal{E}(\mathbf{v}_i, \mathbf{p}) = 
        \forall \mathbf{q} \in \mathcal{Q}({\mathbf{v}_i,\mathbf{p}}):\ 
        \min_{\mathbf{s} \in \mathcal{PC}} \|\mathbf{q} - \mathbf{s}\| \ge \tau$
        \State $S_{\mathbf{v}_i} \gets \frac{1}{|\mathcal{PC}_{\bar{o}}|} 
        \sum_{\mathbf{p} \in \mathcal{PC}_{\bar{o}}} \mathds{1}_{\mathcal{E}(\mathbf{v}_i, \mathbf{p})}$
        \Comment{Visibility ratio}
        \If{$S_{\mathbf{v}_i} >S_{best}$}
            \State $S_{best} \gets S_{\mathbf{v}_i}$, $\mathbf{v}_{best} \gets \mathbf{v}_i$
        \EndIf
    \EndFor
    \State \Return $\mathbf{v}_{best}$ \Comment{Best viewpoint maximizing visibility}
\end{algorithmic}
\end{algorithm}


\section{Experiment}
\subsection{Experimental Setting}
\paragraph{Datasets and Evaluation Metrics.} We evaluate our proposed approach on two established goal-oriented navigation benchmarks: 1) \textit{GOAT-Bench}~\cite{goatbench} (Multi-modal lifelong open-vocabulary dataset, 360 episodes, 36 scenes, 2669 total subtasks, 36 novel goal categories). 2) \textit{HM3D-ObjNav} (HM3D-Semantics-v0.2~\cite{hm3d}  from 2023 Habitat Challenge, 1000 episodes, 36 scenes, 6 goal categories). To validate the generalization capability of our method to novel environments, all evaluations on GOAT-Bench are conducted on the \textit{Val Unseen} split. Following standard practice, we assess navigation performance using Success Rate ($\text{SR} = \frac{N_{\text{success}}}{N_{\text{total}}}$) and Success weighted by Path Length ($\mathrm{SPL} = \frac{1}{N_{\text{total}}} \sum_{i=1}^{N_{\text{total}}} S_i \frac{l^{s}_i}{\max(l^{s}_i, l^{a}_i)}$), where $S_i \in \{0,1\}$ denotes binary task success, and $l^{s}_i$ and $l^{a}_i$ represent the shortest path distance and the actual path traversed by the agent.

\paragraph{Implementation Details.} We used the official success distance threshold from the benchmark: 0.25 for GoatBench and 1.0 for HM3D-ObjNav. We employ GPT-4o (2024-08-06) as our primary Vision-Language Model (VLM) via the OpenAI API. Additional analyses utilizing Qwen-VL-Max as an alternative VLM backbone, along with extended evaluations on the HM3D-OVON benchmark~\cite{ovon}, are deferred to the supplementary material. 

\subsection{Main Experimental Results}
We will show the main comparison results with other state-of-the-art methods on the Goat-Bench \cite{goatbench} and HM3D-ObjNav \cite{hm3d} benchmark.  

\subsubsection{Goat-Bench Benchmark}

\noindent The results in Table \ref{table:goat} also show the outstanding performance of our \texttt{MSGNav} in tackling multimodal goals, lifelong navigation task settings. \texttt{MSGNav} achieves the best performance with 52.0\% SPL and 29.6\% SPL compared to both previously training-based and train-free methods. It surpasses the previous state-of-the-art training-based method MTU3D \cite{mtu3d} by 4.8\% SR and 2.1\% SPL, demonstrating its superior navigation capabilities.  These results highlight the effectiveness of our multi-modal scene graph in tackling multi-modal lifelong navigation tasks.

\begin{table}[!t]
\centering
\renewcommand\arraystretch{1.2} 
\resizebox{\linewidth}{!}{
\begin{tabular}{cccc}
\toprule

\textbf{Method}         & \textbf{Training-free} & \textbf{SR} $\uparrow$ & \textbf{SPL} $\uparrow$  \\ \hline
SenseAct-NN Monolitic  \cite{goatbench}            & ×                      & 12.3                  & 6.8           \\
Modular CLIP on Wheels \cite{goatbench}            & \checkmark                      & 16.1                  & 10.4          \\
Modular GOAT   \cite{goatbench}                  & \checkmark                      & 24.9                  & 17.2          \\
SenseAct-NN Skill Chain  \cite{goatbench}          & ×                      & 29.5                  & 11.3          \\
VLMnav     \cite{vlmnav}       & \checkmark                      & 20.1                  & 9.6           \\
DyNaVLM     \cite{dynavlm}          & \checkmark                      & 25.5                  & 10.2          \\
3D-Mem$^\dag$   \cite{3dmem}              & \checkmark                      & 28.8                  & 15.8          \\
TANGO    \cite{tango}                & \checkmark                      & 32.1                  & 16.5          \\
MTU3D         \cite{mtu3d}                   & ×                      & \uline{47.2}                  & \uline{27.7}          \\ \midrule
\texttt{MSGNav} (Ours)                        & \checkmark                      & \textbf{52.0}         & \textbf{29.6} \\ \bottomrule

\end{tabular}
}
\caption{Experiments on the ``\textit{Val Unseen}'' split of GOAT-Bench. ``\dag'' denotes the results we reproduced due to different settings.}
\label{table:goat}
\end{table}

\subsubsection{HM3D-ObjNav Benchmark}
 Table \ref{table:goat} demonstrates the superior zero-shot multi-modal open-vocabulary navigation capabilities of our \texttt{MSGNav}, we further evaluate it on the widely used HM3D-ObjNav benchmark. As shown in Table \ref{table:hm3d}, \texttt{MSGNav} achieves a state-of-the-art Success Rate (SR) of 74.1\%, which is 1.9\% higher than that of the previous best-performing method WMNav \cite{Wmnav}, and significantly outperforms other prior methods. Although our Success Path Length (SPL) is nearly the same as WMNav without any significant advantage, this may be because the VVD module prioritizes viewpoints with a wide field of view over the absolute shortest path.

\begin{table}[!t]
\centering

\renewcommand\arraystretch{1.1} 
\begin{tabular}{cccc}
\toprule
\textbf{Method} &  \textbf{Training-free} & \textbf{SR} $\uparrow$& \textbf{SPL} $\uparrow$\\ \hline
L3MVN  \cite{L3mvn}                    & \checkmark                      &36.3 &15.7          \\

SG-Nav \cite{sg-nav}                & \checkmark                      & 49.6                 & 25.5         \\

InstructNav   \cite{Instructnav}                    &  \checkmark                     & 58.0                  & 20.9 \\
CompassNav  \cite{Compassnav}                    & \checkmark                      & 59.6                  & 26.9         \\ 
Schrodinger'sNav \cite{SchrodingersNavigator} & \checkmark                      & 60.9                  & 23.7 \\ 
VLFM      \cite{vlfm}                & ×                      & 62.6                 & 31.0         \\
DORAEMON   \cite{doraemon}                    &  \checkmark                      & 66.5                  & 20.6 \\ 

WMNav  \cite{Wmnav}                     & \checkmark                      & 72.2                  & 33.3         \\
\midrule
\texttt{MSGNav} (Ours)                    & \checkmark                      & \textbf{74.1}         & \textbf{33.4}  \\ \bottomrule
\end{tabular}                      
\caption{Experiments on the HM3D-ObjNav benchmark.}
\label{table:hm3d}
\end{table}

\begin{table*}[!t]
\centering
\resizebox{1\linewidth}{!}{
\begin{tabular}{cccc|cccc|cccc}
\bottomrule
\multicolumn{4}{c|}{\textbf{Module}}                        & \multicolumn{4}{c|}{\textbf{Success Rate}}                         & \multicolumn{4}{c}{\textbf{SPL}}                                       \\ \cline{5-12}
M3DSG & VVD & AVU & CLR & Overall       & Category      & Language      & Image         & Overall       & Category      & Language      & Image         \\ \hline
\multicolumn{4}{c|}{ }                                 & 28.8          & 29.9          & 26.8          & 29.5          & 20.2          & 21.0          & 18.7          & 21.0          \\
\checkmark              &              &              &              & 43.8          & 56.0          & 35.1          & 40.2          & 28.0          & 31.3          & 23.5          & 29.1          \\
\checkmark              & \checkmark            &              &              & 56.3          & 61.6          & 55.0          & 52.3          & 34.7          & 34.9          & 33.9          & 35.2          \\
\checkmark              & \checkmark            & \checkmark            &              & 55.3          & 58.6          & 51.7          & 55.7          & 36.7          & \textbf{35.3} & \textbf{34.8} & 40.1          \\
\checkmark              & \checkmark            &              & \checkmark            & 53.2          & \textbf{64.6} & 38.5          & 55.7          & 32.9          & 34.8          & 25.8          & 37.7          \\ 
\checkmark              & \checkmark            & \checkmark            & \checkmark            & \textbf{60.0} & 63.6          & \textbf{57.2} & \textbf{59.1} & \textbf{37.0} & 35.0          & 33.4          & \textbf{42.6} \\ \toprule
\end{tabular}
}
\caption{Component ablation experiment across the first episode of each scene on the ``\textit{Val Unseen}'' split of GOAT-Bench. The first row without any module, which represents our baseline model 3D-Mem \cite{3dmem} results. ``VVD'', ``AVU'', and ``CRV'' represent the Visibility-based Viewpoint Decision module, Adaptive Vocabulary Update module, and Closed-loop Reasoning and Verification module.}
\label{table:abla}
\end{table*}

\begin{table*}[!t]
\setlength{\tabcolsep}{1.2mm}
\centering
\begin{tabular}{c|cccc|cccc}
\bottomrule
\multirow{2}{*}{\textbf{3D Scene Graph}}                                                              & \multicolumn{4}{c|}{\textbf{Success Rate}}                                     & \multicolumn{4}{c}{\textbf{SPL}}                                          \\ \cline{2-5} \cline{6-9} & Overall & Category & Language & Image & Overall & Category & Language & Image \\ \hline
Node-only                   & 51.8             & \textbf{63.6}              & 44.0              & 47.8           & 31.2             &\textbf{39.3}              & 20.5              & 33.8           \\
            Traditional graph \cite{Conceptgraphs}             & 56.2             & \textbf{63.6}              & 52.8              & 52.3           & 32.7             & 35.3              & 27.3              & 35.5           \\ M3DSG (Ours)               & \textbf{60.0}             & \textbf{63.6}              & \textbf{57.2}              & \textbf{59.1}           & \textbf{37.0 }            & 35.0              &\textbf{33.4}              & \textbf{42.6}          \\ \toprule
\end{tabular}
\caption{Scene graph experiment across the first episode of each scene on the ``\textit{Val Unseen}'' split of GOAT-Bench. ``Node-only''  indicates Concept-graph \cite{Conceptgraphs} without object relation edges. ``Traditional graph'' indicates Concept-graph \cite{Conceptgraphs}.}
\label{table:graph}
\end{table*}

\subsection{Ablation Analysis}
In this section, we further analyze the effectiveness and advancement of each module in the \texttt{MSGNav} system. Specifically, this includes component ablation, the advantages of multimodal edges, and demonstrating how the VVD module aids in “\textit{last-mile}” decision-making.
\subsubsection{Effective of Component}
Table \ref{table:abla} demonstrates the impact of each component. Firstly, introducing MSG3D yielded an improvement of 15.0\% in SR and 7.8\% in SPL (row 2). Subsequently, VVD further gains 12.5\% in SR and 6.7\% in SPL (row 3). Notably, introducing either AVU (row 4) or CLR (row 5) alone can only improve a part of categories. This is because AVU's new vocabulary may introduce some insufficient perception results, while the strict decision-making of CLR may hinder navigation toward detailed language-described targets. However, the optimal performance of simultaneously incorporating AVU and CLR (row 6) shows that  AVU and CLR are complementary. AVU can supplement extra perception information for the strict decision-making of CLR, while CLR can address the insufficient perception of AVU. These results validate the effectiveness of each component.

\subsubsection{Advantage of M3DSG}
Table \ref{table:graph} shows the experimental results comparing MSGNav using Node‑only, Concept‑graph \cite{Conceptgraphs}, and our M3DSG. Concept‑graph outperforms Node‑only by 4.4\% in SR and 1.5\% in SPL, especially for Language and Image goals, underscoring the value of relationship edges. M3DSG further boosts these categories with a total of 3.8\% SR and 4.3\% SPL, showing its effectiveness in scene understanding. It is worth noting that node-only performs well on category goals, mainly because such tasks rely on concise object sets rather than complex contextual information.

\begin{table}[!t]
\setlength{\tabcolsep}{1.8mm}
\centering
\begin{tabular}{c|cc}
\bottomrule
\textbf{Success Threshold}&\multicolumn{2}{c}{\textbf{Success Rate} (\texttt{MSGNav})}\\ \cline{2-3}
 $d $  (m) &   w/o VVD   & w/ VVD\\ \hline
0.25 (standard)              & 33.91                 & 51.97          \\
0.35                & 48.93               & 59.69          \\
0.45                & 54.59                & 61.73          \\
0.55              & 57.44                & 63.03          \\
0.65              & 59.65                & 63.78          \\
0.75               & 60.40              & 64.52          \\
0.85               & 61.30               & 65.22          \\
1.00              & 62.38  & 66.52          \\ \toprule
\end{tabular}
\caption{Experiments on the ``\textit{Val Unseen}'' split of GOAT-Bench with various success thresholds. VVD module represents the Visibility-based Viewpoint Decision module. }
\label{table:thre}
\end{table}

\subsubsection{Decision-making for ``Last-mile''}
\begin{figure}[!t]
    \centering
    \includegraphics[width=0.85\columnwidth]{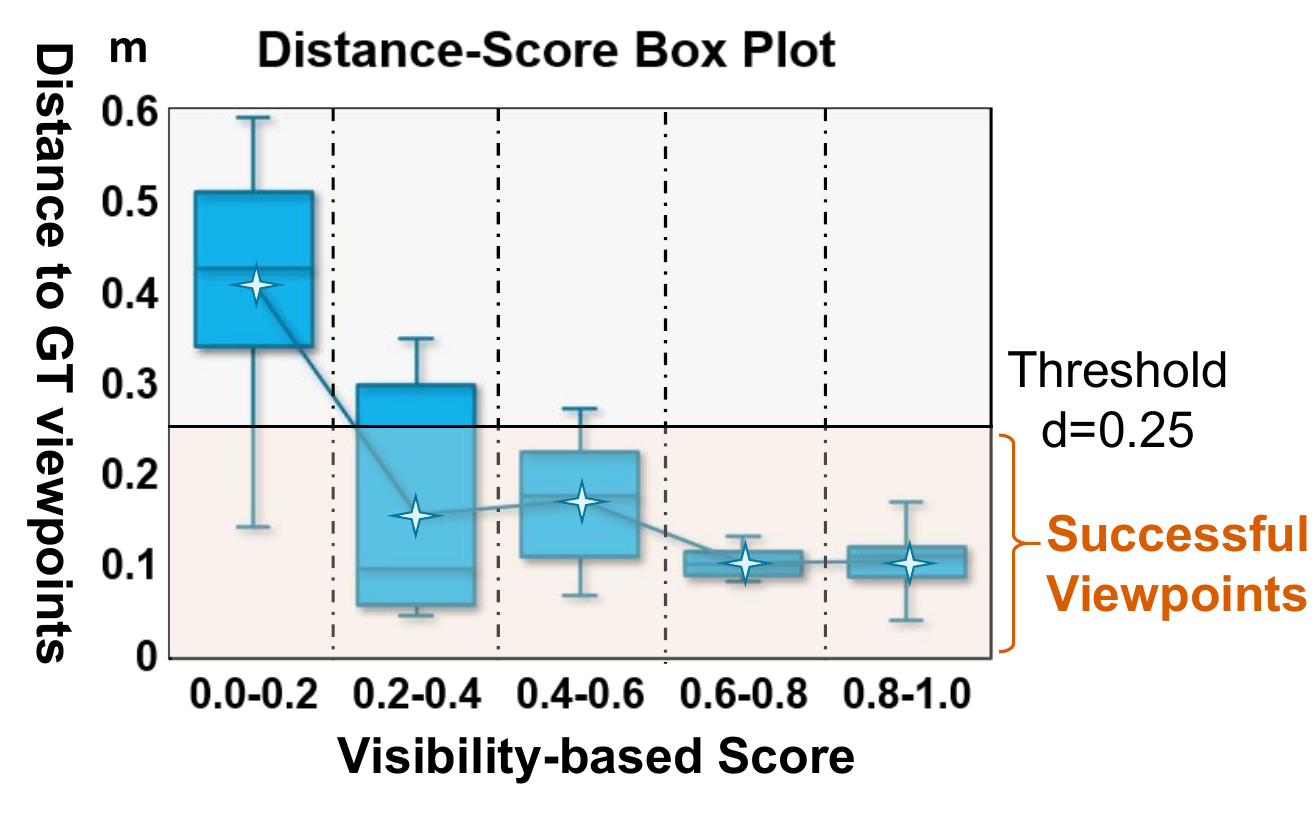}
  \caption{Statistical box plot of candidate viewpoint scores computed by the VVD module and distances from GT viewpoints.}
  \label{distance-score}
\end{figure}

Table~\ref{table:thre} validates the effectiveness of VVD in addressing the ``last‑mile'' problem. Navigating without VVD yields only a 33.91\% SR at the standard threshold of $d$ = 0.25\,m. As the threshold is relaxed, its SR can be significantly improved, indicating that lots of failed tasks stop near the target due to incorrect navigation point selection. Navigating with VVD can improve SR to 51.91\%, which primarily recovers cases within 0.25–1\,m. Fig.~\ref{distance-score} further confirms the reliability of the visibility score estimated by the VVD: the higher-scoring viewpoint is closer to the GT viewpoint, and viewpoints with scores above 0.6 are always sufficiently close to the GT viewpoints. Nevertheless, Table~\ref{table:thre} also shows that the gains of navigating with VVD can still be achieved as the threshold is relaxed, demonstrating that VVD mitigates but does not eliminate the ``last‑mile'' problem.


\section{Conclusion}
\label{sec:conclusion}

In this paper, we propose the \texttt{MSGNav}, a zero‑shot embodied navigation framework built upon a Multi-modal 3D Scene Graph (M3DSG) that preserves visual information for efficient construction, robust perception, and unconstrained vocabulary. Meanwhile, we integrate a visibility‑based viewpoint decision module in \texttt{MSGNav} to resolve the last‑mile problem. Extensive experiments on GOAT‑Bench and HM3D‑ObjNav datasets demonstrate consistent and notable gains over existing methods, highlighting the effectiveness of the proposed multi‑modal scene representations in embodied navigation.

\noindent \textbf{Limitations and future work.} Despite these advantages of \texttt{MSGNav}, scene graph-based methods still face low inference efficiency due to the latency of VFMs and VLMs, suggesting future work on faster graph construction and inference for real‑time deployment. Additionally, while the “last-mile” problem has been mitigated by the VVD module, it is not fully resolved. Exploring reinforcement learning approaches, e.g. active perception, to further solve this challenge is also a way worthy of attention.

\noindent  \textbf{Acknowledgements.} This work was supported in part by the National Natural Science Foundation of China (No. 42571514) and the Zhongguancun Academy (Grant No. 02012405).
\small
{

    \bibliographystyle{ieeenat_fullname}
    \bibliography{main}
}

\clearpage
\setcounter{page}{1}
\maketitlesupplementary

\renewcommand{\thefootnote}{\arabic{footnote}}

\section{Implement Details}
This section outlines the implementation details of our approach, including computational resource, hyper-parameters, and the vision foundation models (VFMs) used for constructing M3DSG.
\subsection{Computational Resource}
Our approach primarily employs a Vision-Language Model (VLM) for inference. Consequently, when utilizing API calls to closed-source models, it places only modest demands on local GPU computational resources. All experiments were conducted using NVIDIA A800-SXM4-40GB graphics cards, with each card requiring approximately 8 GB to 10 GB of VRAM to perform VFMs inference. This configuration ensures stable performance while maintaining cost-effectiveness in resource usage.

Meanwhile, inference with open-source large models such as GPT-OSS \footnote{GPT-OSS: \url{https://github.com/openai/gpt-oss}}  and LLAVA  \footnote{LLAVA: \url{https://github.com/haotian-liu/LLaVA}}  is also feasible. However, this configuration demands additional GPU memory that scales in proportion to the parameter size of open-source large models.

\subsection{Hyper-parameters}
The details of the hyperparameters of \texttt{MSGNav}  are shown in Table \ref{table:hyp}. Additionally, we maintain consistency with 3D-Mem and Conceptgraph for other parameters in the 3D scene graph construction.

\begin{table}[!h]
\setlength{\tabcolsep}{1.8mm}
\centering
\begin{tabular}{c c}
\toprule
\textbf{Hyperparameters}&\textbf{Value} \\ \midrule
Radii set: $\mathbf{R}$                          & [0.5, 0.75]  (m)        \\
Camera height: $h$,                           & 1.5 (m)          \\
Success threshold: $d$              & 0.25 (m)          \\

Samples number: $K$                         & 20          \\
Adjacency distance: $\theta$ & 3.5 (m) \\
Obstruction distance: $\tau$                       & 0.05 (m)          \\

Top‑k related objects: $|\textbf{O}^{rel}|$                      & 20          \\

\midrule
Hfov of camera                     & 120$^{\circ}$         \\
Image resolution              & 1280x1280        \\
Image prompt resolution           & 512x512 \\
\bottomrule
\end{tabular}
\caption{The details of hyperparameters of \texttt{MSGNav}. }
\label{table:hyp}
\end{table}

\subsection{VFMs for Constructing M3DSG}
As outlined in the main paper, the 3D scene reconstruction pipeline primarily incorporates the following Visual Feature Modules (VFMs). First, YOLO-World \cite{yolo} is applied for object detection and room recognition. Next, SAM \cite{sam} is utilized to generate semantic masks. Finally, CLIP \cite{clip} is employed to extract visual feature representations of the objects.

\begin{figure}[!t]
    \centering
    \includegraphics[width=1\columnwidth]{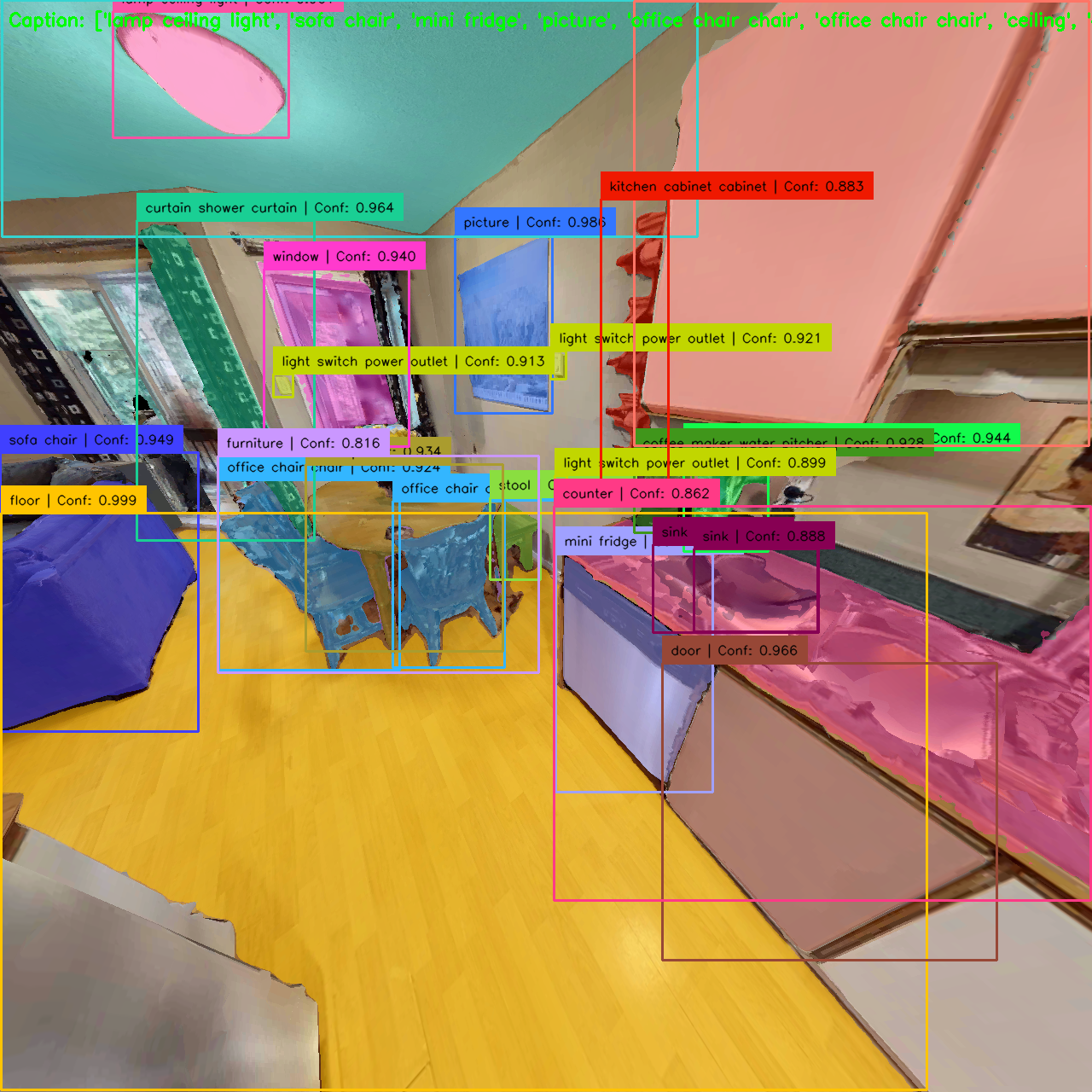}
  \caption{A visualization example of perception results during constructing the M3DSG.}
  \label{Vis_per}
\end{figure}

An example of visualized perception results is shown as Fig. \ref{Vis_per}, illustrating the outputs obtained after processing with YOLO-World and SAM for open-vocabulary object detection and semantic segmentation.

And the detailed information of VFMs is shown in Table \ref{table:vfm}.

\begin{table}[!h]
\setlength{\tabcolsep}{1.8mm}
\centering
\resizebox{1\linewidth}{!}{
\begin{tabular}{c c c}
\toprule
\textbf{VFM}&\textbf{Version}&\textbf{Size} \\\midrule
Yolo-World \cite{yolo} & YOLOv8x-world	 & 141.11 MB\\
SAM \cite{sam} & SAM-Large & 1.16 GB\\
CLIP \cite{clip}& ViT-H-14-quickgelu/dfn5b &12.20 GB \\
\bottomrule
\end{tabular}
}
\caption{The detailed information of VFMs for constructing M3DSG. }
\label{table:vfm}
\end{table}

\begin{figure*}[t]
  \centering
  \includegraphics[width=\linewidth]{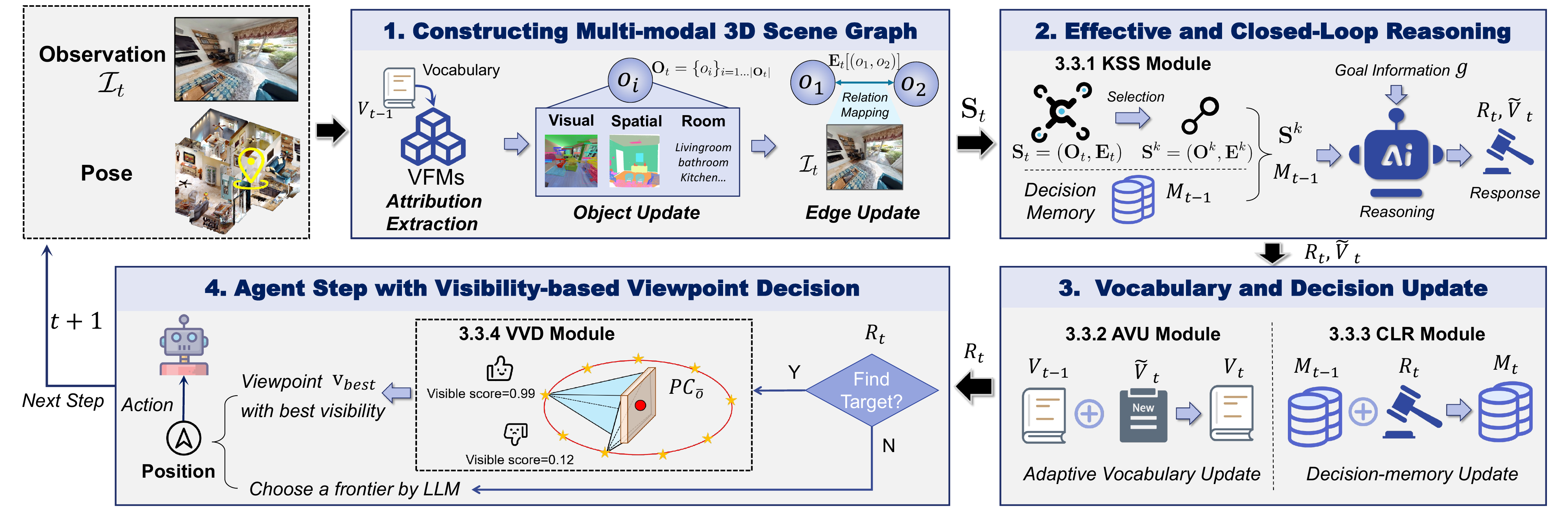}
  \caption{
    \textbf{Detailed end-to-end processing flow of the proposed framework.} The system executes a continuous sense-reason-act loop across four key stages: (1) multi-modal 3D scene graph construction via Vision Foundation Models (VFMs), (2) closed-loop reasoning based on the extracted sub-graph $\mathbf{S}^k$ and historical memory, (3) adaptive updates to the semantic vocabulary and decision memory, and (4) visibility-based viewpoint decision or frontier exploration. Please refer to the corresponding text for a detailed description of each mathematical symbol and data flow.
  }
  \label{fig:e2eframework}
\end{figure*}

\begin{table}[!t]
\centering
\renewcommand\arraystretch{1.1} 
\begin{tabular}{cccc}
\toprule
\textbf{Method} &  \textbf{Training-free} & \textbf{SR} $\uparrow$& \textbf{SPL} $\uparrow$\\ \hline
BCRL \cite{dagger}                        & $\times$                       & 20.2                      & 8.2           \\
DAgger \cite{dagger}                    & $\times$                       & 18.1                   & 9.4           \\
DAgRL \cite{dagger}                 & $\times$                       & 41.3                   & 21.2          \\
RL \cite{RL}                       & $\times$                       & 39.2                   & 18.7           \\
VLFM \cite{vlfm}                      & \checkmark                      & 35.2                   & 18.6          \\
Uni-NaVid \cite{uninavid}                & $\times$                       & 41.3                   & 21.2          \\
DAgRL+OD \cite{ovon}              & $\times$                       & 35.8                   & 21.2          \\
MTU3D \cite{mtu3d}                    & $\times$                       & \textbf{55.0}                  & \uline{23.6}          \\ \midrule
\texttt{MSGNav} (Ours)                    & \checkmark                      & \uline{48.3}         & \textbf{27.0}  \\ \bottomrule
\end{tabular}                       
\caption{Comparison of navigation performance on the \textbf{Val Seen} split of HM3D-OVON. This experiment evaluates the model's optimization effectiveness and navigation efficiency within the training distribution.}
\label{table:ovon_seen}
\end{table}

\subsection{Illustration of End-to-end Framework}
In addition to the framework diagram in the main paper, we also provide a more intuitive end-to-end framework diagram that systematically illustrates the data flow and processing logic between modules, as shown in Fig. \ref{fig:e2eframework}. 

The system operates in a continuous sense-reason-act loop at each timestep $t$, which can be detailed in four key stages: \textbf{(1) Constructing Multi-modal 3D Scene Graph:} Given the current observation $\mathcal{I}_t$ and agent pose, Vision Foundation Models (VFMs) extract visual, spatial, and room-level attributes to incrementally update the object nodes and relational edges of the scene graph $\mathbf{S}_t$. \textbf{(2) Effective and Closed-Loop Reasoning:} A Key Sub-graph Selection (KSS) module filters $\mathbf{S}_t$ into a task-relevant sub-graph $\mathbf{S}^k$. An LLM agent then reasons over $\mathbf{S}^k$, historical decision memory $M_{t-1}$, and goal information $\mathcal{G}$ to generate an action response $R_t$ and proposed vocabulary updates $\tilde{V}_t$. \textbf{(3) Vocabulary and Decision Update:} The Adaptive Vocabulary Update (AVU) module integrates $\tilde{V}_t$ to form the current vocabulary $V_t$, while the Closed-Loop Reasoning (CLR) module records $R_t$ to update the decision memory $M_t$. \textbf{(4) Agent Step with Visibility-based Viewpoint Decision:} If the target is identified, the VVD module selects the optimal viewpoint $\mathbf{v}_{best}$ that maximizes the visibility score of the target point cloud $PC_{\bar{o}}$; otherwise, the LLM selects a new frontier for exploration in step $t+1$.

\begin{table*}[!t]
    \centering
    \begin{tabular}{ccccccccc}
        \toprule
        \textbf{VLM} &  \multicolumn{2}{c}{\textbf{Overall} (2669)} & \multicolumn{2}{c}{{\textbf{Object Category} (991)}} & \multicolumn{2}{c}{\textbf{Language} (856)} & \multicolumn{2}{c}{\textbf{Image} (822)}\\
        \cmidrule(r){2-3} \cmidrule(r){4-5} \cmidrule(r){6-7} \cmidrule(r){8-9} 
         (\texttt{MSGNav}) & Success Rate & SPL & Success Rate & SPL & Success Rate & SPL & Success Rate & SPL \\ \midrule
         GPT-4o \cite{gpt4o}  & 51.97 & 29.56  & 53.38  &29.98  & \textbf{47.43} & 25.08 & 54.99 &\textbf{33.71} \\
         Qwen-VL-Max \cite{qwen}  & \textbf{52.79} & \textbf{30.79}  & \textbf{56.10}   & \textbf{31.53} & 45.56 & \textbf{27.18}  & \textbf{56.33} &33.67 \\
        \bottomrule
    \end{tabular}
    \caption{Experiments of our \texttt{MSGNav} method using different VLMs on the ``\textit{Val Unseen}'' split of GOAT-Bench. The number following each category represents the sample size.}
    \label{tab:detail-goat-bench}
    \vspace{-10pt}
\end{table*}

\section{More Experimental Results}
\subsection{Additional Analysis on HM3D-OVON}

To further investigate the foundational perception and optimization stability of the proposed \texttt{MSGNav}, we conduct supplementary experiments on the HM3D-OVON dataset \cite{ovon}. For this analysis, we specifically report the results on the ``\textit{Val Seen}'' split to evaluate the model's performance when navigating within familiar environment distributions, which serves as a benchmark for the model's upper-bound efficiency and basic semantic mapping capabilities.

\noindent As presented in Table \ref{table:ovon_seen}, we compare \texttt{MSGNav} with several state-of-the-art training-based and training-free methods on the seen environments. While the training-based method MTU3D \cite{mtu3d} achieves a higher Success Rate (SR) of 55.0\% by leveraging extensive environment-specific learning, \texttt{MSGNav} significantly outperforms all methods in terms of Success weighted by Path Length (SPL), reaching \textbf{27.0\%}. 

This result is particularly noteworthy as \texttt{MSGNav} is a \textit{training-free} approach. The superior SPL indicates that even in seen scenarios, our method generates more efficient and purposeful navigation paths compared to models that might overfit to specific trajectories. These supplementary results confirm that \texttt{MSGNav} maintains robust basic navigation logic and high path efficiency, which provides a solid foundation for the zero-shot generalization capabilities demonstrated in our primary experiments (Table 1 in the main paper).

\subsection{Quantitative experiments on Goat-Bench}
We present the detailed experimental results in Table 2 of the main paper, showing the detailed performance of \texttt{MSGNav} across different categories with various VLM. As shown in the table \ref{tab:detail-goat-bench}, \texttt{MSGNav} demonstrates strong adaptability across different VLMs instead of relying on the specific model. \texttt{MSGNav} achieves comparable results on both state-of-the-art VLMs, Qwen-VL-Max and GPT-4o. It is worth noting that the results on Qwen-VL-Max are slightly better than those on GPT-4o. This is primarily due to the costly token fees of GPT-4o, our prompts were primarily implemented based on Qwen-VL-Max, and they may be better suited for Qwen-VL-Max. However, the excellent results achieved on GPT-4o without any prompt adjustments demonstrate that the system does not overly rely on prompt engineering.

\section{Analysis of the Exploration}

\subsection{Additional Details of VVD Module}

\begin{figure*}[!t]
\centering
    \includegraphics[width=0.8\textwidth]{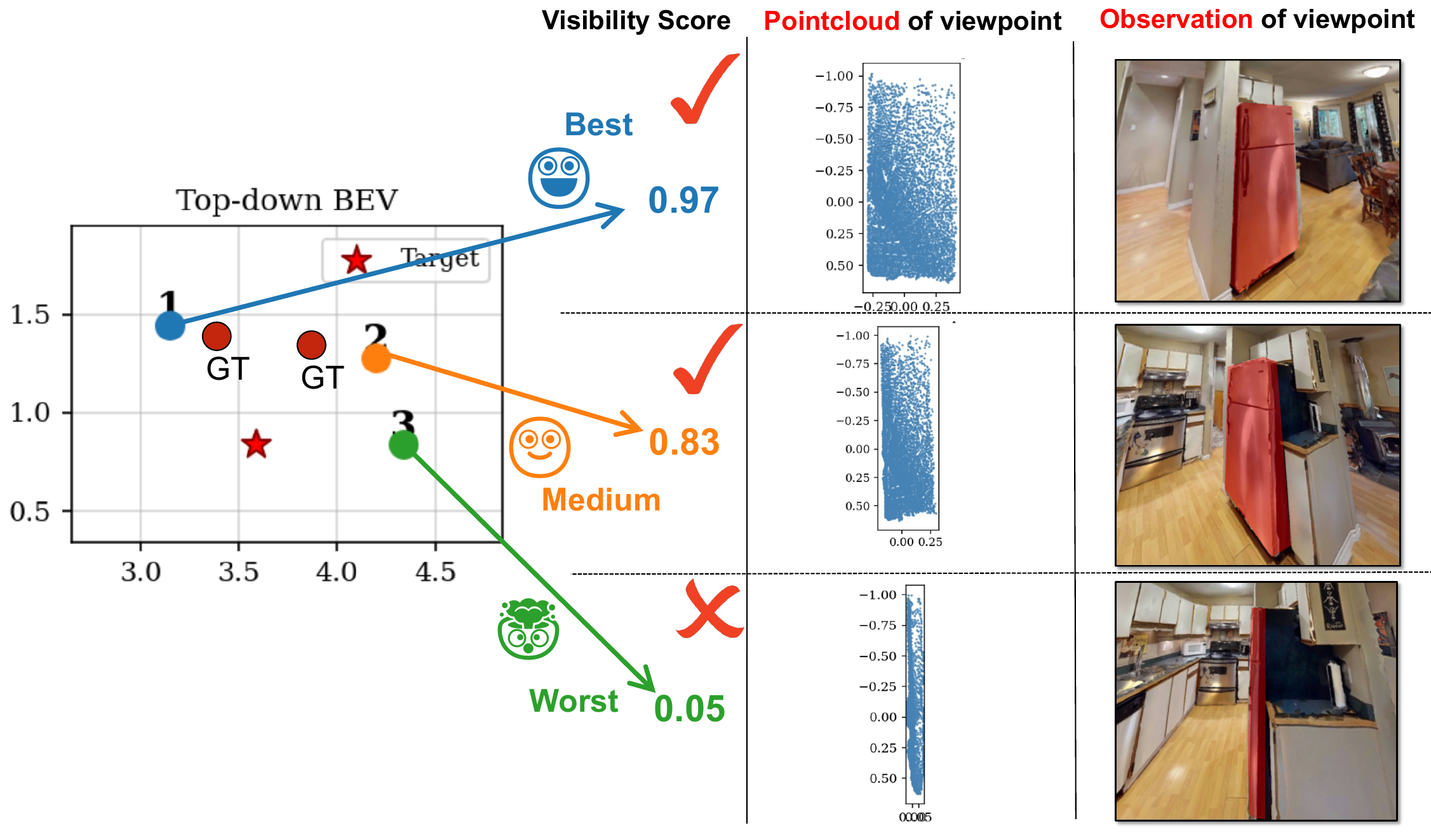}
  \caption{\textbf{Qualitative Visualization of the Visibility-based Viewpoint Decision (VVD) Module.} The central plot shows a top-down Bird's-Eye View (BEV) of the agent's spatial environment. Red stars indicate potential targets, while red circles mark the Ground Truth (GT) target viewpoints. The VVD module evaluates candidate viewpoints (numbered 1, 2, and 3), calculating a visibility score based on ray-casting to the target point cloud. Viewpoint 1 achieves the 'Best' visibility score (0.97), and its corresponding first-person observation provides a clear, unoccluded view of the red refrigerator. Viewpoint 2 receives a 'Medium' score (0.83), capturing a partial, side-angle view. Viewpoint 3 receives the 'Worst' score (0.05), as its line of sight is heavily occluded by other environmental structures.}
  \label{vis_vvd_more}
\end{figure*}

We provide an example illustrating how the VVD module selects optimal viewpoints. As shown in Fig. \ref{vis_vvd_more}, candidate viewpoints located closer to the ground-truth (GT) viewpoints receive higher visibility scores. In this example, the highest-scoring viewpoint (Viewpoint 1, score 0.97) is positioned directly in front of the target, offering an unobstructed field of view that clearly captures the target object.

By contrast, the medium-scoring viewpoint (Viewpoint 2, score 0.83) is positioned on the side of the target, thereby capturing only partial visual information. Notably, Viewpoint 3 (score 0.05) is severely occluded by environmental structures, which may be less discernible purely from a simplified 2D bird’s-eye view (BEV) representation. In such cases, the VVD module correctly identifies the occlusion by computing 3D line-of-sight, consistently assigning low visibility scores to these obstructed viewpoints.

To further clarify the mechanics of the VVD module, we expand upon the operations detailed in Algorithm 2 of the main paper. The algorithm systematically evaluates line-of-sight occlusion in 3D space to identify the optimal navigation endpoint.

\begin{enumerate}
    \item \textbf{Candidate Generation (Lines 1-4):} The algorithm first computes the centroid $\mathbf{c}_{\bar{o}}$ of the localized target's point cloud $\mathcal{PC}_{\bar{o}}$. It then generates a set of $K$ candidate viewpoints in a concentric circle (or multiple concentric circles defined by radii set $\mathbf{R}$) around this centroid. Crucially, it filters this set to retain only viewpoints $\mathbf{V}_c$ that reside in traversable free space, ensuring the agent can physically reach them.
    \item \textbf{Ray Construction (Line 7):} For a given candidate viewpoint $\mathbf{v}_i$ and a specific point $\mathbf{p}$ belonging to the target object, the algorithm mathematically defines a ray segment $\mathcal{Q}(\mathbf{v}_i,\mathbf{p})$. This segment connects the viewpoint and the target point. To account for the physical volume of the agent and potential sensor noise, the parameter $\tau$ introduces a safety margin at both ends of the ray.
    \item \textbf{Occlusion Evaluation (Line 8):} The core visibility condition $\mathcal{E}(\mathbf{v}_i, \mathbf{p})$ evaluates whether the defined ray $\mathcal{Q}$ is obstructed by the broader scene point cloud $\mathcal{PC}$. A target point $\mathbf{p}$ is considered visible if and only if the minimum distance between any point $\mathbf{q}$ on the ray and any point $\mathbf{s}$ in the environment point cloud is strictly greater than or equal to the obstruction margin $\tau$.
    \item \textbf{Visibility Scoring (Lines 9-13):} The algorithm computes an aggregate visibility score $S_{\mathbf{v}_i}$ for the candidate viewpoint. This score represents the ratio of target points $\mathbf{p} \in \mathcal{PC}_{\bar{o}}$ that satisfy the visibility condition $\mathcal{E}$. The viewpoint achieving the highest ratio is selected as the optimal navigation target $\mathbf{v}_{best}$, guaranteeing the agent finishes its trajectory with a robust, unoccluded view of the target.
\end{enumerate}

\subsection{Visualization about VVD Module}
\begin{figure}[!t]
    \centering
    \includegraphics[width=0.8\columnwidth]{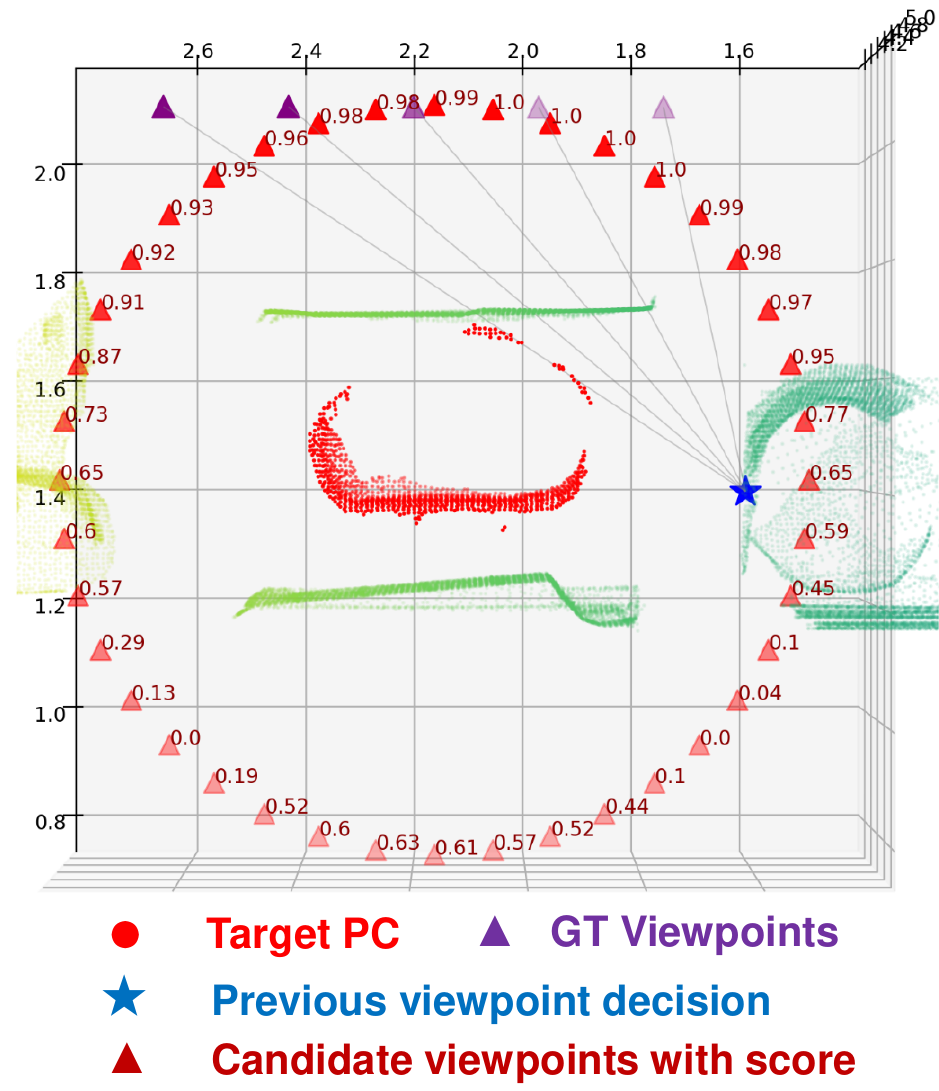}
  \caption{An example of viewpoint decision by the VVD module. This shows the BEV view of the point cloud in the scene and the corresponding viewpoint information.}
  \label{vis_vvd}
\end{figure}

\begin{figure*}[!t]
    \centering
    \includegraphics[width=\textwidth]{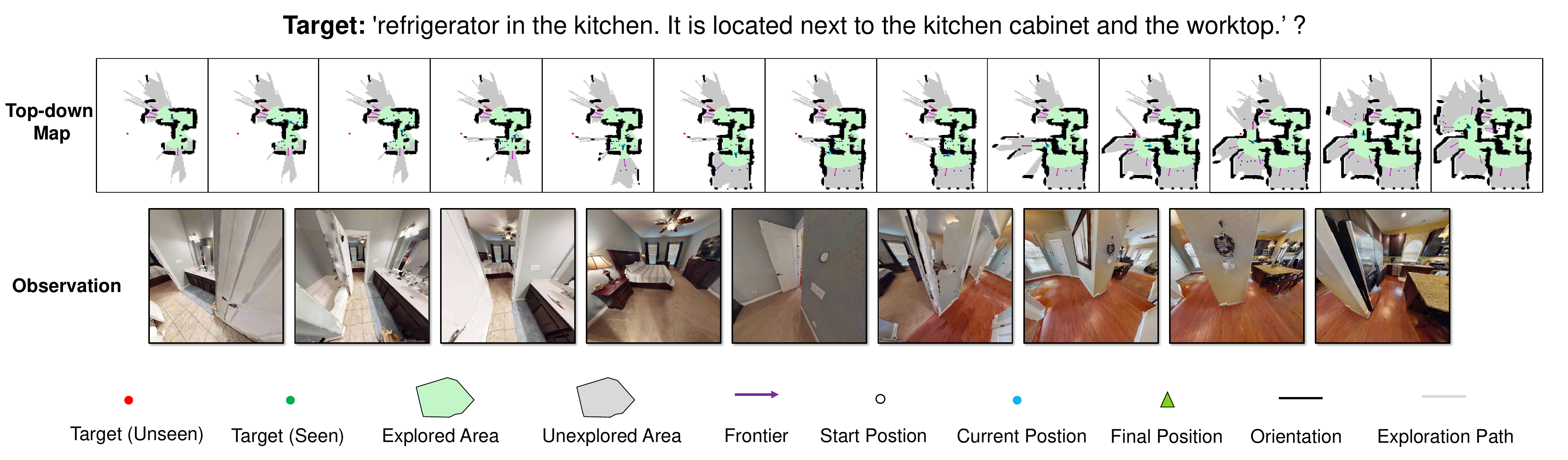}
  \caption{This visualization illustrates the process of an agent exploring and navigating to a target described by natural language—“refrigerator in the kitchen. It is located next to the kitchen cabinet and the worktop”—in an indoor environment. It is composed of three parts from top to bottom: the Top-down Map sequence, the Observation sequence (first-person visual inputs), and the legend, which respectively display the global exploration state, the agent’s local visual perception, and the meaning of each visual element.}
  \label{explore_vis}
\end{figure*}

\begin{figure*}[!h]
\centering
    \includegraphics[width=1\textwidth]{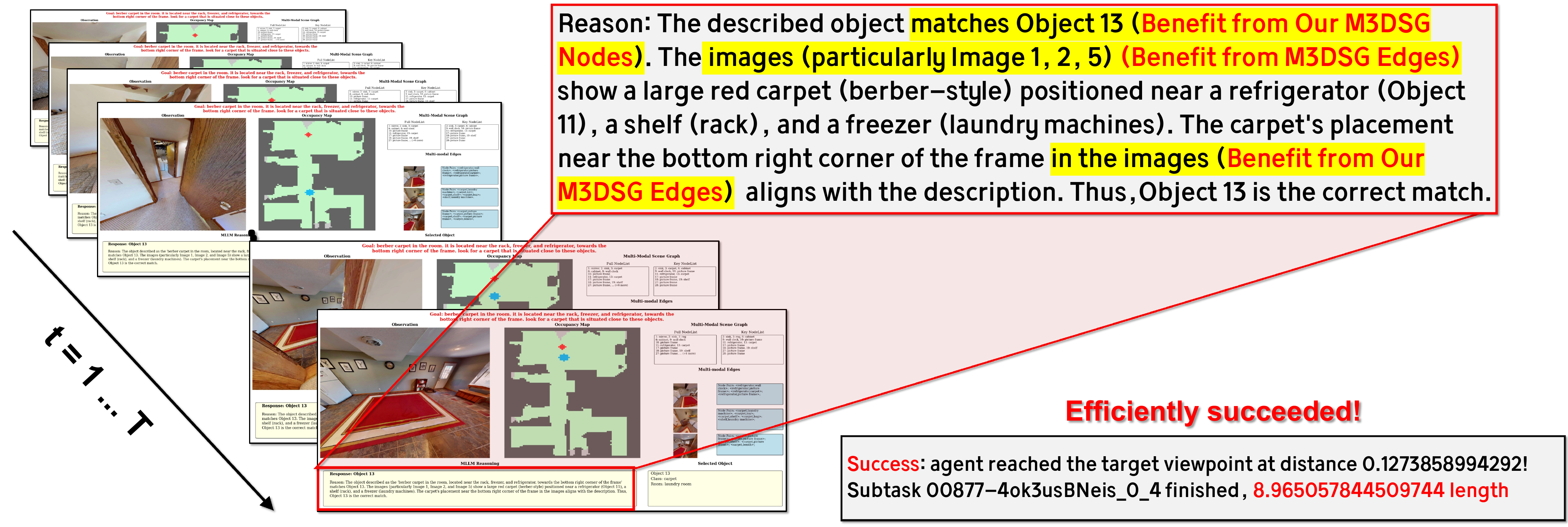}
    \caption{\textbf{Visualization of the VLM reasoning process based on M3DSG.} The agent successfully locates the target "berber carpet" by leveraging M3DSG Nodes to match the object candidates (Object 13) and Multi-modal Edges (images) to verify complex spatial relationships, such as the carpet's placement near the refrigerator (Object 11) and its position in the bottom right corner of the frame. This accurate grounding leads to an efficiently successful navigation task.}
  \label{fig:vis_sg}
\end{figure*}

We provide a real-world example illustrating how the VVD module selects optimal viewpoints. As shown in Fig. \ref{vis_vvd}, candidate viewpoints located closer to the ground-truth (GT) viewpoints, marked by the purple triangle, receive higher visibility scores. In this example, the highest-scoring viewpoint is almost coincident with the GT viewpoint and is positioned directly in front of the target (top), offering an unobstructed field of view.

By contrast, the previous viewpoint selection was farther from the GT viewpoint and positioned on the side of the target, thereby capturing only partial visual information. Notably, no GT viewpoint exists at the rear (lower side) of the target due to occlusion caused by point cloud data (green dots) at the same height within the scene, which may be less discernible in the bird’s-eye view (BEV) representation. In such cases, the VVD module consistently assigns low visibility scores (0.1–0.6) to these occluded viewpoints.

\subsection{Visualization about Exploration}
As shown in Fig. \ref{explore_vis}, we present the natural language-guided target navigation by visualizing the agent’s process of exploring and localizing the target described as “refrigerator in the kitchen. It is located next to the kitchen cabinet and the worktop” to intuitively demonstrate its closed-loop reasoning. 

The agent starts at the predefined Start Position and begins to explore adjacent areas.  At this stage, the target refrigerator is still unseen (indicated by a red dot). The agent continuously moves along the Frontier (the boundary between explored and unexplored areas, marked by purple arrows) to uncover more regions. Its movement trajectory is recorded as the Exploration path (light gray line), and its Current Position (blue dot) is updated dynamically as it navigates through different rooms. As the agent ventures into the kitchen area, the Observation sequence captures the visual input of the refrigerator adjacent to the kitchen cabinet and worktop. Simultaneously, in the Top-down Map, the target marker changes from a red dot (Target (unseen)) to a green dot (Target (seen)), and the agent reaches the Final Position (green triangle), completing this hard navigation task.

This visualization comprehensively demonstrates how the agent uses the multi-modal scene graph to sequentially reason and localize the target object in an unknown environment. With the help of our multi-modal scene graph, agents can gradually explore and locate targets even when they are extremely distant. This further validates the effectiveness of our multi-modal scene graph for reasoning in zero-shot embodied navigation tasks.

Furthermore, to explicitly illustrate the critical role of our proposed M3DSG in the granular decision-making process, we visualize the Vision-Language Model (VLM) reasoning process in Fig. \ref{fig:vis_sg}. In this episode, the agent is tasked with finding a target based on a complex spatial description: "herber carpet is the room. It is located near the rack, freezer, and refrigerator towards the bottom right corner of the frame". As the exploration proceeds ($t=1 \dots T$), the M3DSG incrementally updates. When matching the target, the VLM relies on the M3DSG Nodes to identify the candidate, explicitly noting that the described object matches Object 13. More importantly, it heavily depends on the Multi-modal Edges (stored images) to verify fine-grained spatial relations. The reasoning process highlights that specific images (particularly Image 1, 2, and 5) show a large red carpet positioned near a refrigerator (Object 11), a shelf, and a freezer. Additionally, the visual edges confirm the carpet's placement near the bottom right corner of the frame, perfectly aligning with the language description. By combining node attributes and visual edge evidence, the agent confidently concludes that Object 13 is the correct match. Ultimately, this accurate multimodal reasoning allows the agent to efficiently succeed in the task, reaching the target viewpoint at a remarkably close distance of 0.127m.

\begin{figure}[!b]
    \centering
    \includegraphics[width=\columnwidth]{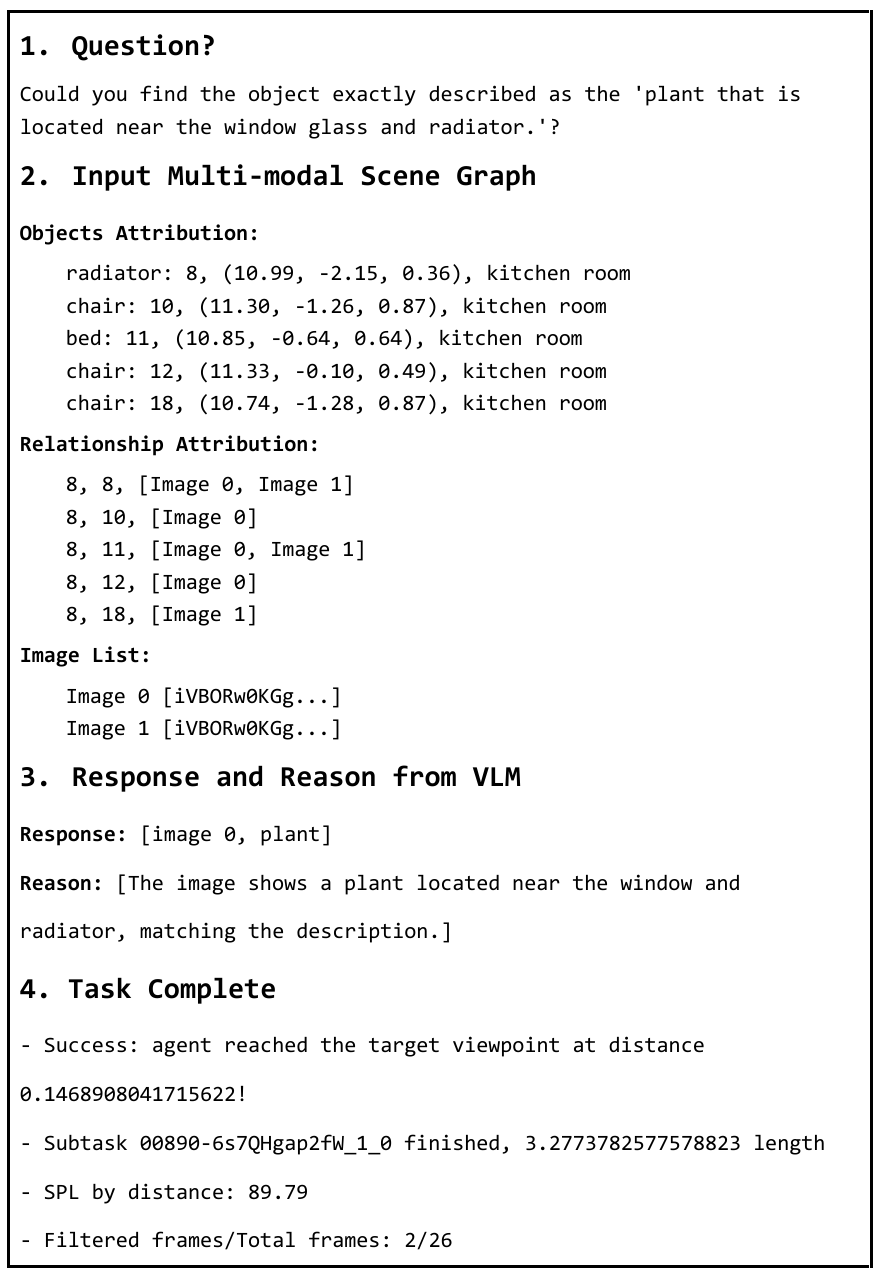}
  \caption{Example illustrating the use of a Vision-Language Model (VLM) for multimodal scene graph reasoning.}
  \label{pana}
\end{figure}

\subsection{Analysis of Reasoning based on M3DSG }
We present an example illustrating the use of a Vision-Language Model (VLM) for multimodal scene graph reasoning. As shown in Fig. \ref{pana}, when given the query “plant that is located near the window glass and radiator”, the conventional textual description in the scene graph failed to include information pertinent to the target. In contrast, the VLM, utilizing the supplementary visual information embedded in the scene graph, successfully identified the target matching the query. The model then enriched the scene graph through vocabulary supplementation and updates. As a result, the task was completed with a Success weighted by Path Length (SPL) of 89.79, requiring only about 3 m of exploration length. This demonstrates that retaining images within the multimodal scene graph substantially improves exploration efficiency.

\begin{figure}[!t]
    \centering
    \includegraphics[width=\columnwidth]{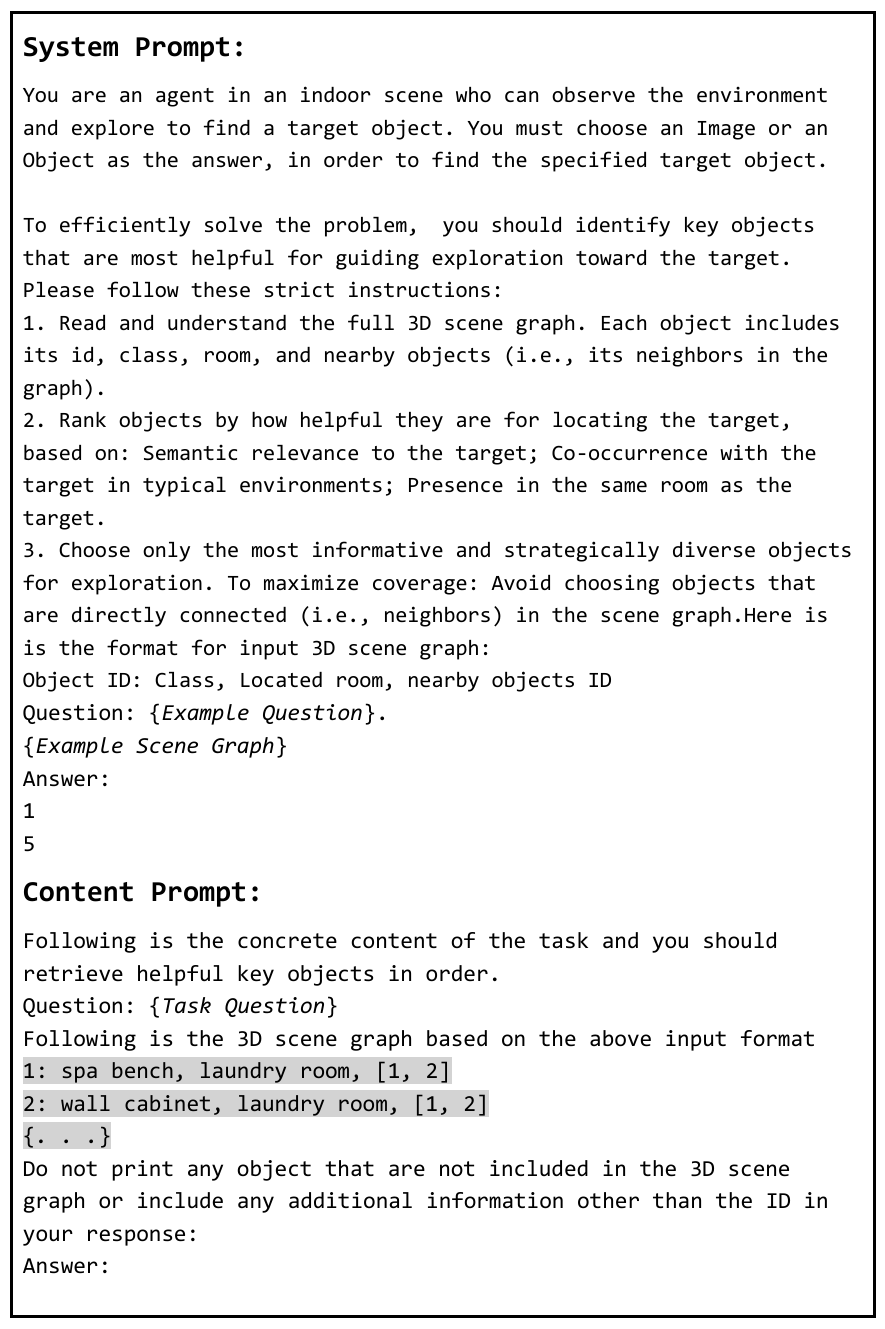}
  \caption{Prompt 1 for ``top-k object nodes selection in KSS module". The placeholders \{...\} will be replaced by the corresponding information. The gray-highlighted text represents the information of compressed M3DSG.}
  \label{p1}
\end{figure}

\section{VLM Reasoning and Prompt}
\begin{figure*}[!t]
    \centering
    \includegraphics[width=0.75\textwidth]{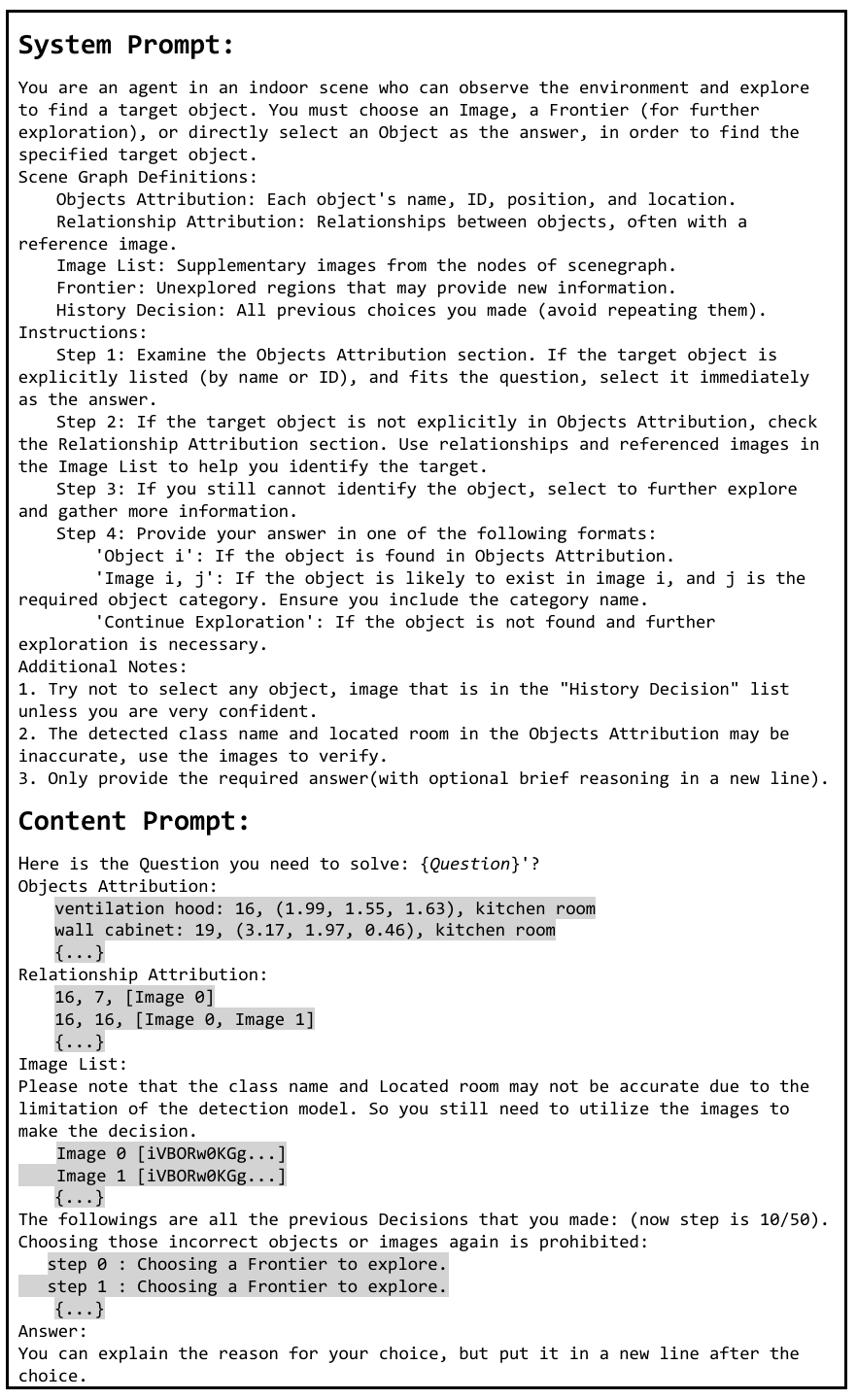}
  \caption{Prompt 2 for ``\textit{find target in exploration inference}". The placeholders \{...\} will be replaced by the corresponding information. The gray-highlighted text represents the information of the full M3DSG.}
  \label{p2}
\end{figure*}

\begin{figure*}[!t]
    \centering
    \includegraphics[width=0.65\textwidth]{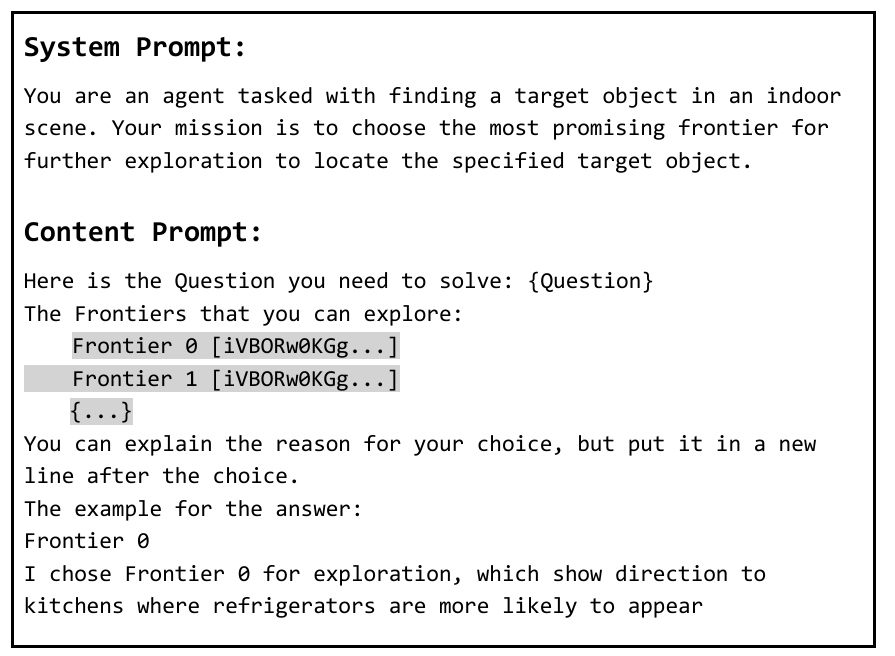}
  \caption{Prompt 3 for ``\textit{explore frontier in exploration inference}". The placeholders \{...\} will be replaced by the corresponding information. The gray-highlighted text represents the information of frontier images as in 3D-Mem \cite{3dmem}.}
  \label{p3}
\end{figure*}

\begin{figure*}[!t]
    \centering
    \includegraphics[width=0.65\textwidth]{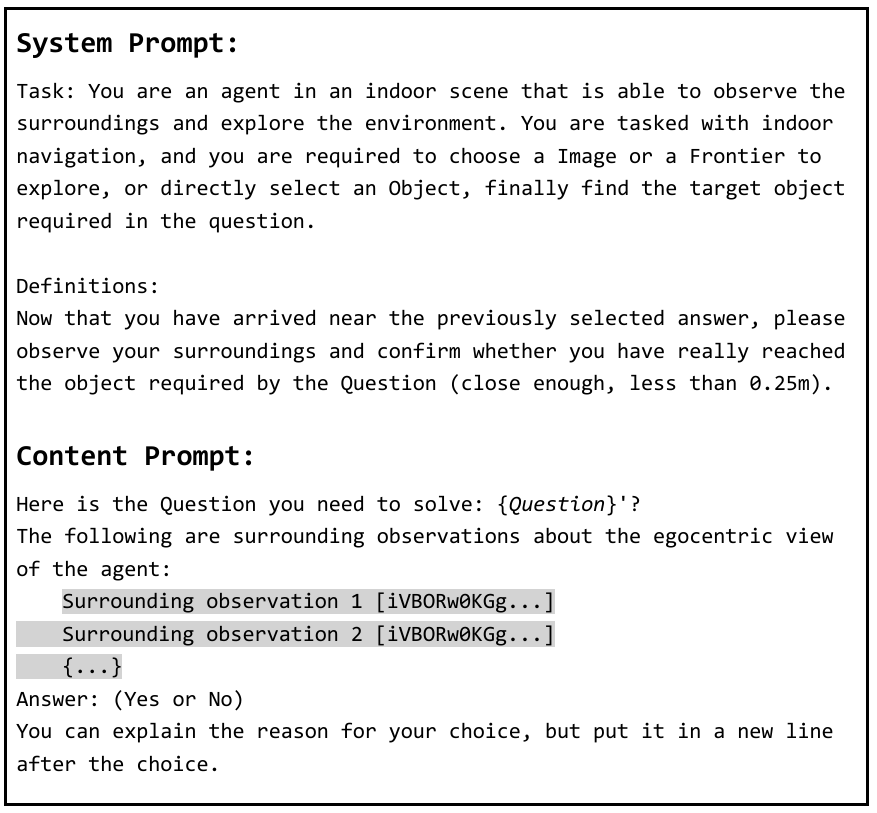}
  \caption{Prompt 4 for ``\textit{task completion verification}". The placeholders \{...\} will be replaced by the corresponding information. The gray-highlighted text represents the information of observation images from the past steps.}
  \label{p4}
\end{figure*}

Our \textbf{MGSNav} method leverages Vision-Language Model (VLM) reasoning in three scenarios during the navigation process:

\begin{itemize}
    \item \textbf{Top-k object nodes selection in KSS} (As shown in Fig. \ref{p1}) — Within the Key Scene Selection (KSS) module, the VLM is applied to infer the top-$k$ target-related nodes from the compressed scene graph.
    
    \item \textbf{Exploration inference}  — The KSS-compressed key scene subgraph is processed by the VLM to identify either the target node or the exploration frontier. This reasoning is performed in two sequential steps:
    \begin{enumerate}
        \item \textbf{Find target}. If the target is present in the 3D scene graph, input the scene graph information to identify the target ID. (As shown in Fig. \ref{p2}) 
        \item \textbf{Explore frontier}. If absent, input the frontier image to determine the exploration frontier image ID. (As shown in Fig. \ref{p3}) 
    \end{enumerate} 
    This decomposition facilitates more accurate VLM reasoning.
    
    \item \textbf{Task completion verification} — Once the agent considers the task completed, the VLM utilizes observation images from the past steps to perform a final inference, thereby validating task credibility. (As shown in Fig. \ref{p4})
\end{itemize}

To better understand the process of VLM reasoning, we present the prompts used for the three VLM reasoning scenarios described above.

\end{document}